\documentclass{article}

\usepackage{natbib}

\usepackage{multirow}
\usepackage[hidelinks]{hyperref}
\usepackage{graphicx}
\usepackage{xspace}
\usepackage{changepage}
\usepackage{arydshln}
\usepackage{tabularx} 
\usepackage{amsmath}
\usepackage{placeins} 
\usepackage{booktabs}
\usepackage{subcaption}
\usepackage{multicol}
\usepackage[preprint]{template}
\usepackage[utf8]{inputenc} 
\usepackage[T1]{fontenc}    
\usepackage{url}            
\usepackage{setspace}
\usepackage{colortbl}
\usepackage{longtable}
\usepackage[table]{xcolor}
\definecolor{darkgreen}{HTML}{3cb44b}
\definecolor{darkred}{HTML}{ff7256}
\definecolor{xdcolor}{HTML}{19b5af}
\usepackage{amsfonts}       
\usepackage{nicefrac}       
\usepackage{algorithm}
\usepackage[noend]{algpseudocode}

\title{Universality of Layer-Level Entropy-Weighted Quantization Beyond Model Architecture and Size}

\author{
\hspace{0.8em}Alireza Behtash
\hspace{0.8em}Marijan Fofonjka
\hspace{0.8em}Ethan Baird\\
\hspace{0.8em}{\bf Tyler Mauer}
\hspace{0.8em}{\bf Hossein Moghimifam}
\hspace{0.8em}{\bf David Stout}
\hspace{0.8em}{\bf Joel Dennison} \\ \\
webAI \\ \\
\texttt{\{alireza.behtash,marijan.fofonjka,ethan\}@webai.com}
\\
\texttt{\{tyler,hossein,David,joel.dennison\}@webai.com}
}
\date{}

\begin{document}
\maketitle

\begin{abstract}
We present a novel approach to selective model quantization that transcends the limitations of architecture-specific and size-dependent compression methods for Large Language Models (LLMs) using Entropy-Weighted Quantization (EWQ). By analyzing the entropy distribution across transformer blocks, EWQ determines which blocks can be safely quantized without causing significant performance degradation, independent of model architecture or size. Our method outperforms uniform quantization approaches, maintaining Massive Multitask Language Understanding (MMLU) accuracy scores within 0.5\% of unquantized models while reducing memory usage by up to 18\%. We demonstrate the effectiveness of EWQ across multiple architectures—from 1.6B to 70B parameters—showcasing consistent improvements in the quality-compression trade-off regardless of model scale or architectural design. A surprising finding of EWQ is its ability to reduce perplexity compared to unquantized models, suggesting the presence of beneficial regularization through selective precision reduction. This improvement holds across different model families, indicating a fundamental relationship between layer-level entropy and optimal precision requirements. Additionally, we introduce FastEWQ, a rapid method for entropy distribution analysis that eliminates the need for loading model weights. This technique leverages universal characteristics of entropy distribution that persist across various architectures and scales, enabling near-instantaneous quantization decisions while maintaining 80\% classification accuracy with full entropy analysis. Our results demonstrate that effective quantization strategies can be developed independently of specific architectural choices or model sizes, opening new possibilities for efficient LLM deployment.
\end{abstract}

\section{Introduction}

The widespread adoption of LLMs has been constrained by their substantial computational and memory requirements, particularly as model sizes continue to grow exponentially \citep{brown2020language, chowdhery2022palm, hoffmann2022training}. As LLMs become increasingly integral to various applications, from natural language processing to automated reasoning, the need for efficient deployment solutions becomes more critical. A typical LLM with 7-70B parameters requires 14-140GB of memory in full precision, making deployment challenging even in well-resourced data centers and practically impossible on edge devices or consumer hardware.

The memory bottleneck is particularly acute when serving multiple users or handling long context windows. For instance, the key-value cache required for processing long sequences can consume several gigabytes of additional memory per request. This challenge is compounded in production environments where multiple model instances must run concurrently to handle user traffic, leading to substantial infrastructure costs and deployment complexity.

Neural network quantization has emerged as a promising approach to address these deployment challenges by reducing resource usage and accelerating inference through lowering the bit precision of model parameters \citep{choi2018pact, hubara2021accurate, yao2022zeroquant, park2022nuqmm, gholami2022survey}. We can define quantization as a way of reducing data precision from typical 32-bit or $\text{bfloat16}$ to smaller bits ($<=$ 8-bit), which in turn lowers the size of the model and speeds up matrix multiplications involved in the attention mechanism. While quantization offers significant memory savings—potentially reducing model size by 75\% or more—maintaining model performance under reduced precision remains a fundamental challenge.

Quantization techniques can be broadly categorized into {\it uniform precision quantization} and {\it mixed precision quantization}. While uniform precision quantization is widely applied to reduce the size of transformer layers in LLMs, its indiscriminate application often leads to significant performance degradation. This degradation occurs because different layers in transformer-based models exhibit varying sensitivity to quantization, necessitating more nuanced approaches \citep{Wang2018, Cai2020, zadeh2020gobo, Ganesh_2021}. For example, early layers processing raw input tokens and final layers producing output logits typically require higher precision than intermediate layers.

Recent research has focused on addressing the challenges posed by outlier activations, which represent a key impediment to effective uniform low-precision quantization. Mixed-precision quantization has shown promise in mitigating this issue by maintaining outlier channels at higher precision \citep{dettmers2022gpt3, zhao2024atom, ashkboos2023quik, zhao2024atom}. Another emerging approach is {\it invariant random rotation}, which aims to suppress outliers and enable more effective uniform low-precision quantization \citep{ashkboos2024quarot, liu2024spinquant, wei2023outlier}. While both methods improve the signal-to-quantization noise ratio and reduce quantization errors locally, they have yet to demonstrate substantial performance advantages over 16-bit precision models. For instance, {\it SpinQuant} \citep{liu2024spinquant} applied at 4-bit precision to Llama-3-8b \citep{llama3meta} shows approximately 20\% higher perplexity compared to the 16-bit baseline, despite significant optimization efforts.

Alternative approaches such as {\it SmoothQuant} \citep{xiao2024smoothquant} and {\it Activation-Aware Quantization (AWQ)} \citep{yuan2023asvd, lin2024awq} have been developed to enable effective 8-bit quantization of both weights and activations. These methods employ sophisticated techniques, including offline migration of quantization difficulty from activations to weights and channel equalization. However, they typically require access to the entire model for activation distribution analysis, making them impractical in resource-constrained environments \citep{kim2023squeezellm}. This is why weight-only quantization represents a more suitable use case for model compression, exemplified by methods like {\it GPTQ} \citep{frantar2022gptq} or {\it FineQuant} \citep{kim2023finequant}. GPTQ converts quantized weights to float16 during inference for matrix multiplication operations and FineQuant uses a fine-grained quantization algorithm that incorporates group-wise quantization and adaptive selection of granularity. While these approaches can achieve performance gains through reduced data loading with minimal accuracy loss, they present practical limitations in production environments specially in distributed settings. For example, handling long context lengths and batch processing increases the memory footprint of key-value (KV) cache substantially. Or uniform quantization of all transformer blocks happen to increase perplexity significantly. Still, even with these challenges the weight-only quantization seems to be a promising approach for the purposes of this paper. So we would like to pose the following important question.

\noindent {\bf Question:} {\it Is it possible to devise an architecture-agnostic, optimal post-training weight-only quantization method that, given resource constraints, produces an on-the-fly\footnote{By `on-the-fly' we mean \(O(1)\) time complexity.} quantized model that remains competitive with the original-precision model while delivering fast inference and memory efficiency?}

To provide a production-grade answer, we introduce Entropy-Weighted Quantization (EWQ), a systematic framework for selective model compression that preserves performance while substantially reducing memory requirements. Our approach extends the theoretical foundations of information-theoretic neural network compression \citep{park2017, xu2018deep, dong2019hawq} to address the unique challenges of large language models. While entropy-based quantization has proven effective for traditional machine learning architectures \citep{park2017} and has recently shown promise in quantization-aware training of LLMs \citep{shen2024}, its application to runtime weight-only quantization remains unexplored. This gap is particularly significant given that LLMs exhibit markedly heterogeneous entropy distributions across their transformer layers—a characteristic that distinguishes them from smaller neural architectures and necessitates more sophisticated quantization strategies.

Our approach's distinguishing feature is its ability to facilitate efficient on-the-fly quantization, making it particularly well-suited for deployment on consumer-grade hardware with limited resources. Unlike traditional mixed-precision methods that demand significant computational overhead for activation analysis \citep{dettmers2022gpt3, kim2023qmoe}, EWQ's weight-centric approach allows for rapid deployment while maintaining adaptability to different hardware constraints. By intelligently mapping the entropy characteristics of LLM weights to appropriate precision levels, EWQ achieves an optimal balance between model performance and resource efficiency. This ensures that quantized models maintain competitive performance with their full-precision counterparts while significantly reducing memory footprint and accelerating inference speed. The architecture-agnostic nature of EWQ is further demonstrated by the emergence of a universal approximator for the quantization of transformer layers, enabling rapid quantization—termed {\it FastEWQ}—without the need to load weights. 

\section{Background and Related Work}
\subsection{Model Quantization}

In more technical terms, model (neural network) quantization refers to the process of reducing the numerical precision of weights and activations from 32-bit floating point to lower bit-width representations (typically 8-bit or 4-bit), quantization achieves significant reductions in memory footprint and computational requirements while preserving model functionality \citep{choi2018pact, hubara2021accurate}. Modern quantization approaches can be broadly categorized into two paradigms:

\begin{itemize} \item \textbf{Uniform quantization}: Applies identical precision reduction across all model components, enabling straightforward implementation but often resulting in significant accuracy degradation for sensitive layers \citep{yao2022zeroquant}. \item \textbf{Mixed-precision quantization}: Allocates higher precision to critical layers identified through sensitivity analysis, achieving better accuracy preservation at the cost of increased implementation complexity \citep{dettmers2022gpt3, zhao2024atom}. \end{itemize}

Recent advances in post-training quantization (PTQ) have demonstrated particular promise for LLM deployment. \citet{dettmers2022gpt3} introduced layer-wise adaptive mixed precision for GPT-3 models, maintaining 16-bit precision only for outlier-dominated attention heads. \citet{frantar2022gptq} developed a second-order quantization approach that minimizes layer-wise reconstruction errors, enabling 4-bit quantization of LLaMA models with minimal accuracy loss. The BitLinear layer proposed by \citet{ashkboos2023quik} achieves extreme 1.58-bit quantization through entropy-driven logarithmic representations, though with increased computational overhead.

\subsection{Block Sensitivity Analysis}

The transformer architecture's layered structure exhibits significant heterogeneity in quantization sensitivity across blocks. Early work by \citet{devlin2019bert} demonstrated that initial encoder layers in BERT models capture fundamental syntactic features highly sensitive to precision reduction, while deeper layers encode semantic relationships more tolerant of quantization. This phenomenon was formalized by \citet{dong2019hawqv2hessianawaretraceweighted, dong2019hawqhessianawarequantization, shen2019qberthessianbasedultra} through Hessian-based sensitivity analysis, establishing that attention blocks typically require 2-4× higher precision than feedforward layers.

Three key strategies have emerged for leveraging this sensitivity gradient:

\begin{itemize} \item \textbf{Progressive quantization}: Gradually increases quantization intensity from output to input layers, preserving early layer precision \citep{zadeh2020gobo}. \item \textbf{Attention-aware allocation}: Assigns higher precision to query/key matrices than value projections to maintain attention fidelity \citep{passban2021gobo}. \item \textbf{Task-adaptive thresholds}: Dynamically adjusts layer precision based on downstream task gradients \citep{kim2023squeezellm}. \end{itemize}

In particular, \citet{yuan2023asvd} introduced \textit{activation-aware singular value decomposition (ASVD)}, a post-training compression that addresses challenges in low-rank factorization by managing activation outliers through the scaling of the weight matrix based on the activation distributions, enhancing the accuracy of decomposition. Additionally, it employs an iterative calibration process to optimize layer-specific decomposition, considering the varying sensitivity of different LLM layers. Experiments demonstrate that ASVD can compress networks by 10-20\% without compromising performance. Furthermore, by applying low-rank decomposition to projection matrices in the self-attention module, ASVD achieves up to 50\% reduction in KV cache memory requirements without performance degradation. 

These developments underscore the importance of understanding and leveraging block sensitivity in transformer architectures to inform effective compression strategies.

\subsection{Information-Theoretic Approaches} \label{ssec:info_theory}

The relationship between parameter entropy and quantization robustness originates from fundamental rate-distortion theory \citep{cover2006elements}, where entropy establishes theoretical bounds on lossy compression. In deep learning, \citet{park2017} first operationalized this connection by demonstrating that weight matrices with Shannon entropy \(H(W)\le 4\) bits/parameter could withstand 4-bit quantization with less than 1\% accuracy drop. Their analysis revealed that entropy correlates with parameter redundancy—layers learning simpler patterns (e.g., smooth feature detectors) naturally exhibit lower entropy and greater quantization tolerance \citep{jin2021dynast}.

Recent advances extended this framework to activation entropy. By leveraging activation entropy, methods optimally balance computational efficiency and model accuracy for edge deployment. Using 4-/8-bit multipliers, they employ entropy-driven thresholds to assign 8-bit quantization to high-entropy activations and 4-bit to low-entropy regions, maintaining distortion under 0.5 KL divergence. Adaptive 4/8-bit quantization with 4-bit weights achieves superior accuracy compared to static non-power-of-two baselines. The entropy-regularized objective prioritizes high-information activations, improving performance by 1.2–3.4\% across containment ratios (\(\rho=\) 0-100). This enables a 2.37× on-device speedup, bridging the efficiency-accuracy trade-off. Our method advances these foundations while addressing three persistent challenges discussed next.

Calculating the complete entropy data for all \(n\) transformer layers leads to a complexity of \(O(n)\) including the entropy distribution for activations \citep{shen2024}, which becomes prohibitive in billion-parameter models. We mitigate this by focusing solely on weight-only quantization, maintaining \(O(n)\) time and space complexity. Another challenge is architecture-specific sensitivity. Existing thresholds, such as Park's 4-bit boundary \citep{park2017}, fail for heterogeneous architectures like Mixture of Experts (MoE) models. Instead, we derive architecture-agnostic criteria using FastEWQ to generalize across multiple model families.

A crucial step for any post-training quantization is downloading the model weights, which significantly limits access to hard disk resources. Static quantization policies \citep{ashkboos2023quik} cannot adapt to varying cluster resources. Our on-the-fly optimization framework operates in \(O(n)\) time per resource update for EWQ and \(O(1)\) time with FastEWQ, maintaining Pareto-optimal accuracy-efficiency trade-offs \citep{Abdolrashidi2021}. This synthesis enables the first information-theoretic quantization system that simultaneously achieves sublinear time entropy estimation, cross-architecture validity, and real-time adaptation to deployment constraints.

\subsection{Challenges in Uniform Quantization}

Uniform quantization techniques apply the same precision reductions across all layers and blocks of the model, presenting several significant challenges in practice. Different components of a model exhibit varying levels of sensitivity to precision reductions, making uniform approaches particularly problematic. When all layers are quantized uniformly, the performance of sensitive layers often degrades substantially, leading to reduced model accuracy, coherence, and perplexity. While uniform quantization successfully reduces memory usage and model size, these benefits come at the cost of significantly impacting downstream performance metrics. For example, fully quantized models using int8 (8-bit) or nf4 (4-bit) precision frequently demonstrate notable declines in crucial metrics such as MMLU scores. Furthermore, the rigid nature of uniform quantization provides minimal flexibility for optimization based on task-specific or architecture/size-specific requirements, rendering it particularly suboptimal for specialized applications such as question-answering (QA) or natural language inference.

\subsection{The Case for Mixed Quantization}

\begin{figure}[H]
\centering
\includegraphics[width=1\linewidth]{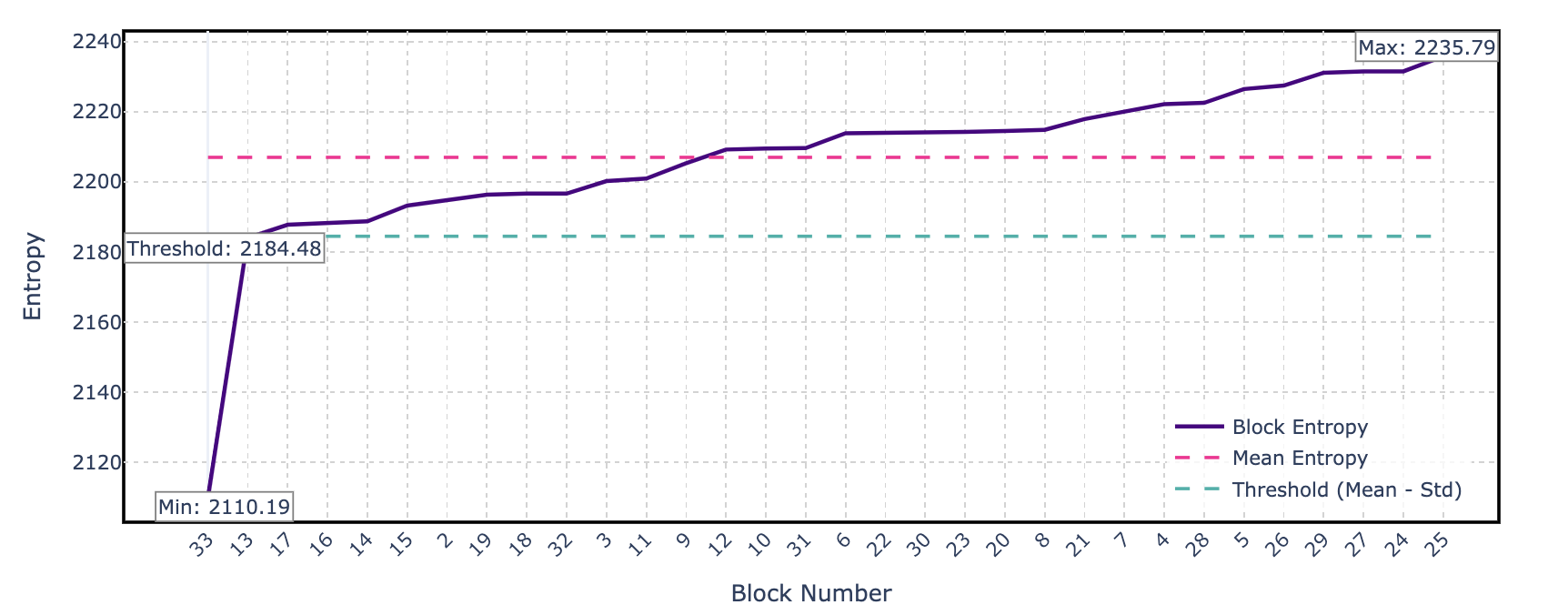}
\caption{Entropy distribution of \texttt{Meta-Llama-3.1-8B-Instruct} model weights with block number. The optimal quantization requires those blocks with lower entropy to be quantized first.}
\label{fig:block_vs_entropy}
\end{figure}

Our preliminary analysis of entropy distributions across transformer blocks, combined with benchmarking the effects of quantization precision on model performance, has revealed several fundamental insights that challenge existing assumptions about model compression. As depicted in Figure~\ref{fig:block_vs_entropy}, transformer blocks exhibit variability in their entropy values, which has profound implications for quantization strategies. Specifically, lower-entropy blocks demonstrate reduced sensitivity to quantization, whereas high-entropy blocks are crucial for maintaining overall model performance. This phenomenon can be attributed to the natural development of hierarchical information structures within transformer blocks during the training process.

Garnier-Brun et al. (2024) introduced a hierarchical filtering procedure for generative models of sequences on trees, allowing for controlled tuning of positional correlations in the data. Their study provides evidence that vanilla encoder-only transformers can approximate exact inference algorithms when trained on root classification and masked language modeling tasks. They observed that correlations at larger distances, corresponding to increasing layers of the hierarchy, are sequentially incorporated by the network during training. This suggests that transformer models inherently develop hierarchical structures, with some layers capturing local information (lower entropy) and others capturing more complex, global information (higher entropy).

Our analysis indicates that the entropy distribution does not follow a universal pattern. Crucially, a layer's entropy—regardless of its position—relates to its handling of global versus local information. Understanding this relationship is important for developing effective quantization strategies with selective precision that preserve model performance while reducing computational complexity. 

Our initial motivation was backed by utilizing the \texttt{Tonic Validate} library to conduct accuracy benchmarks on QA datasets, including MMLU, with various quantization configurations. Mixed precision approaches, utilizing 8-bit precision for 60\% of blocks and 4-bit for the remaining 40\% randomly, achieve the highest answer-similarity at 52\% while maintaining competitive answer consistency at 22\%. Full 8-bit quantization shows improved answer consistency at 26\% but demonstrates lower similarity metrics, suggesting that a mixed quantization can edge out a higher precision global quantization. Complete 4-bit quantization performs poorest across all metrics, emphasizing its unsuitability for precision-demanding tasks. Table~\ref{tab:accuracy_benchmarks} shows a summary of these comparisons.

\begin{table}[H]
\centering
\setlength{\arrayrulewidth}{0.4mm}
\small 
\renewcommand{\arraystretch}{1.2} 
\begin{tabularx}{\linewidth}{|X|c|c|X|}
\hline
\textbf{Configuration}                & \textbf{Similarity} & \textbf{Consistency} & \textbf{Remarks}                                                                 \\ \hline
Mixed Precision (8-bit: 60\%, 4-bit: 40\%) & 52\%                       & 22\%                        & High similarity with competitive consistency.                                     \\ \hline
Fully 8-bit Quantization                & \textless 52\%              & 26\%                        & Better consistency but lower similarity than mixed precision.                     \\ \hline
Fully 4-bit Quantization                & <35\%                              & <12\%                      & Poor performance on both metrics; unsuitable for high-precision tasks.           \\ \hline
\end{tabularx}
\vspace{5px}
\caption{Accuracy benchmarks on QA datasets were evaluated using the \texttt{Tonic Validate} library in the initial phase of the project. The early indication of lower perplexity was observed when similarity was highest for mixed precision, which became a motivating factor to pursue mixed quantization further.}
\label{tab:accuracy_benchmarks}
\end{table}

In the context of QA tasks, the \textit{ResQ} method applies mixed-precision quantization to LLMs, demonstrating that such approaches can maintain model performance while reducing computational costs \citep{saxena2024resq}. Similarly, the \textit{SliM-LLM} framework employs a salience-driven mixed-precision quantization scheme, achieving efficient LLMs with high accuracy \citep{huang2025slimllm}. These findings align with our later observations that mixed-precision quantization strategies can effectively balance model efficiency and performance in QA tasks. 

Building upon insights into the role of entropy in an effective neural network quantization and selective precision of transformer layers, we focus on EWQ and its optimized variant, FastEWQ. These methods enhance model efficiency by assigning precision levels to transformer blocks based on their entropy characteristics. Moving forward, our empirical results will be derived using the MMLU benchmark to align with community standards.

\section{Methodology}
In this section, we delve into the methodology for calculating the entropy of a neural network layer's weight matrix.

\subsection{Entropy Analysis}

The entropy calculation for a neural network layer's weight matrix involves three main steps: flattening the weights, applying the softmax function, and computing the entropy using an information-theoretic approach. The mathematical representation for these calculations is as follows.

\subsubsection{Weight Flattening}
Let the weight matrix of a neural network layer be denoted as \( W \). The weights are flattened into a one-dimensional array
\[
   w_{\text{flat}} = \text{Flatten}(W),
\]
where \( w_{\text{flat}} \) denotes the resulting flattened array of weights, and its length \( n \) represents the total number of parameters in the matrix.

\subsubsection{Softmax Normalization}
To transform the flattened weights into a probability distribution, the softmax function is applied
\[
   p_i = \frac{e^{w_{\text{flat}, i}}}{\sum_{j=1}^{n} e^{w_{\text{flat}, j}}}, \quad \text{for } i = 1, \dots, n,
\]
where \( p_i \) is the probability corresponding to the \( i^{\text{th}} \) weight.
\subsubsection{Entropy Calculation}
The entropy of the weight distribution is computed using the following formula
\[
   H = -\sum_{i=1}^{n} p_i \log(p_i + \epsilon),
\]
where \( \epsilon \) is a small constant (e.g., 0.01) added for numerical stability.

\subsection{Block Entropy Calculation}
For a transformer block, which contains multiple weight matrices from linear and embedding layers, the total weighted entropy of the block is computed using the following formula
\[
   H_{\text{block}} = \frac{\sum_{i} |W_i| H(W_i)}{\sum_{i} |W_i|}, \label{block_entropy}
\]
where \( H_{\text{block}} \) is the total entropy of the block, \( H(W_i) \) is the entropy of the \( i^{\text{th}} \) weight, calculated as
    \[
        H(W_i) = -\sum_{j=1}^{n_i} p_{i,j} \log(p_{i,j} + \epsilon),
    \]
where \( p_{i,j} \) represents the normalized probabilities for the \( j^{\text{th}} \) parameter in the \( i^{\text{th}} \) weight matrix, and \( \epsilon \) is a small constant for numerical stability, \( |W_i| \) is the number of parameters (or size) of the \( i^{\text{th}} \) weight matrix, and \( n_i \) is the number of parameters in \( W_i \).

This formulation ensures that larger weights contribute more to the overall block entropy, providing a weighted representation of the block's variability. By incorporating the sizes of weight matrices into the calculation, the approach captures the relative significance of each matrix in the transformer block. 

\subsection{Block Selection Criteria}
Utilizing the block entropy in Eq.~\eqref{block_entropy}, we establish criteria for selecting transformer blocks for quantization. The selection process involves several steps given below.

\subsubsection{Sorting Blocks by Entropy}
After calculating the entropy \( H_{\text{block}} \) for each transformer block, we sort the blocks in ascending order of entropy. This allows us to prioritize lower-entropy blocks for more aggressive quantization while preserving higher-entropy blocks in higher precision formats to maintain model performance. Mathematically, we express this sorting process as
\[
   \text{Sort}(H_{\text{block}_i}) \quad \text{for} \quad i = 1, \dots, N,
\]
where \( N \) is the total number of blocks, and \( H_{\text{block}_i} \) represents the entropy of the \( i^{\text{th}} \) block.

Organizing the blocks in ascending order of entropy allows for processing low-entropy blocks first, which often contain redundant or low-information content, making them suitable for lower-precision quantization. Conversely, high-entropy blocks, essential for model accuracy due to their significant role in complex token relationships and higher-order representations, are maintained at higher precision. We define the sorted sequence as:
\[ H_{\text{block}_1} \leq H_{\text{block}_2} \leq \cdots \leq H_{\text{block}_N} \]
here \( H_{\text{block}}(i) \) denotes the \( i \)-th element in the sorted entropy list. 
\subsubsection{Computing Mean and Standard Deviation}
Next, we compute the mean and standard deviation of the weighted entropy values of all the blocks. Let \( H_{\text{block}_i} \) be the entropy of block \( i \), and \( N \) be the total number of blocks. The mean entropy \( \mu_H \) and standard deviation \( \sigma_H \) are given by
\[
   \mu_H = \frac{1}{N} \sum_{i=1}^{N} H_{\text{block}_i},~~ \sigma_H = \sqrt{\frac{1}{N} \sum_{i=1}^{N} (H_{\text{block}_i} - \mu_H)^2}.
\]

\subsubsection{Entropy Threshold for Quantization}
Using the mean entropy \( \mu_H \) and the standard deviation \( \sigma_H \), we determine the entropy threshold for quantization. The threshold \( T \) is calculated as
\[
   T = \mu_H - X \cdot \sigma_H,
\]
where \( X \) is a floating point number (\( X \geq 0 \)) that determines how aggressively blocks are quantized. By default, \( X = 1 \).

\subsubsection{Quantization Decision}
In the quantization process referred to as \textit{quantization decision}, blocks with entropy values below a specified threshold \( T \) are targeted for more aggressive quantization methods, such as 4-bit or 1.58-bit precision. This approach is based on the assumption that these low-entropy blocks have a minimal impact on the model's overall performance, allowing for reductions in memory usage and computational demands. Conversely, blocks with entropy values exceeding T but remaining below the mean entropy \( \mu_H\) are considered more critical to model performance and are thus quantized less aggressively, typically using an 8-bit representation. The quantization decision for each block \( b_i \) is defined by 
\[
 Q(b_i) = \begin{cases} 
 4\text{-bit or 1.58-bit} & \text{if } H_{\text{block}_i} \leq T, \\
 8\text{-bit} & \text{if } T < H_{\text{block}_i} \leq \mu_H.
 \end{cases}
\]
This approach ensures that blocks contributing less to the model's overall performance (i.e., blocks with lower entropy) are more aggressively quantized, while those with higher entropy (indicating higher variability or importance) are quantized less aggressively to preserve performance. Blocks with entropy above the mean value \( \mu_H\) are initially left unquantized. A detailed explanation of the quantization strategy is provided in Section~\ref{sect_optimized_dist}.

\subsection{Optimized Distribution of LLM transformer Blocks in Deployment Clusters} \label{sect_optimized_dist}

Based on the calculated quantization decision results, we define an optimization algorithm~\ref{algo-ewq} for distributing LLM transformer blocks across the available machines within a deployment cluster. Consider a machine with \( X \) bytes of available memory for loading transformer blocks during inference and \( Y \) bytes of free disk space. Since model weights must be downloaded to load into memory for execution, the resource limit for each machine in the inference cluster can be defined as \( Z = \min(X, Y) \). If the cluster consists of \( N \) machines, the total available resources for model execution in the inference cluster can be expressed as \( R = \sum Z \), being aggregate resource capacity.

When distributing LLM models, the goal of the optimization algorithm is to maximize the utilization of available resources to preserve the model's unquantized accuracy while minimizing network communication latency between machines. The initial step is to check whether the unquantized model can fit within the cluster's resources. Optimization is necessary when the total unquantized model size, \( W \), exceeds the available resources, \( R \), in the cluster.

The process begins with the results obtained from the quantization decision, where transformer block candidates are ordered in an ascending list based on their calculated weighted block entropy, \( H_{\text{block}} \), and preselected using a defined quantization criterion, \( Q(b_i) \). Blocks below the threshold \( T \) are preselected for 4-bit quantization, while blocks with entropy values above \( T \) but below the entropy mean are assigned 8-bit quantization. If the total model size, after applying these quantization settings, fits within the available resources \( R \), we proceed to promote blocks to higher precision.
\begin{algorithm}[H]
\caption{Optimized Distribution of LLM transformer Blocks}
\label{algo-ewq}
\begin{algorithmic}[1]
\Require \( N \): Number of machines in the cluster
\Require \( X_i, Y_i \): Memory and disk space available on machine \( i \) (\( 1 \leq i \leq N \))
\Require \( W \): Total size of the unquantized model
\Require \( H_{\text{block}} \): List of transformer blocks sorted by weighted entropy
\Require \( Q(b_i) \): Quantization criterion for block \( b_i \)
\Require \( T \): Threshold for 4-bit quantization
\Ensure Optimized quantization and distribution of transformer blocks
\State \( Z_i \gets \min(X_i, Y_i) \) for each machine \( i \)
\State \( R \gets \sum Z_i \) \Comment{Total available resources in the cluster}
\If{\( W \leq R \)}
    \State Deploy model unquantized
    \Return
\EndIf
\ForAll{blocks \( b_i \) in \( H_{\text{block}} \)}
    \If{\( H_{\text{block}}[b_i] \leq T \)}
        \State Assign 4-bit quantization to \( b_i \)
    \ElsIf{\( T < H_{\text{block}}[b_i] \leq \text{mean}(H_{\text{block}}) \)}
        \State Assign 8-bit quantization to \( b_i \)
    \Else
        \State Keep \( b_i \) unquantized
    \EndIf
\EndFor
\State Calculate model size \( S \) after initial quantization
\While{\( S > R \)}
    \ForAll{blocks \( b_i \) in descending order of \( H_{\text{block}} \)}
        \If{\( b_i \) is 8-bit and resources allow}
            \State Promote \( b_i \) to unquantized
        \ElsIf{\( b_i \) is 4-bit and resources allow}
            \State Promote \( b_i \) to 8-bit or unquantized
        \EndIf
    \EndFor
    \State Recalculate \( S \)
\EndWhile
\If{\( S > R \) after Step 4}
    \While{\( S > R \)}
        \State Quantize blocks with lowest \( H_{\text{block}} \) to 1.58-bit
        \State Recalculate \( S \)
    \EndWhile
\EndIf
\State Ensure only blocks with minimal \( H_{\text{block}} \) remain at reduced precision
\State Distribute blocks across machines based on \( Z_i \)
\Return Optimized model quantization and distribution
\end{algorithmic}
\end{algorithm}
If the model resulting from the quantization decision does not fit within the available resources, we examine globally quantized models with 4-bit and 8-bit precision. We preselect the model whose total size is below the available resource capacity \( R \) and begin promoting transformer blocks with the highest entropy to unquantized or 8-bit precision until the total model size approaches the resource limit \( R \). In cases where the globally quantized 4-bit model does not fit within \( R \), we evaluate whether further quantization is possible by reducing blocks with the lowest entropy to 1.58-bit precision, ensuring the model fits within the available resources.

For deployment scenarios with severe resource constraints, such as edge devices or mobile phones, we can adapt this methodology to employ a 4-3bit (or even 2-bit \citep{chee2024quip2bitquantizationlarge}). combination instead of the standard 8-4bit approach. In this configuration, high-entropy blocks are preserved at 4-bit precision while lower-entropy blocks are further compressed to 3-bit precision. Our experiments with this configuration demonstrate that for models deployed on devices with less than 2GB of available RAM, the 4-3bit combination can reduce the model footprint by an additional 18-25\% compared to uniform 4-bit quantization while maintaining acceptable accuracy degradation of less than 5\% on standard benchmarks.
Furthermore, the block distribution algorithm dynamically adjusts to network topology, prioritizing block placement that minimizes cross-machine communication during inference. This is particularly important for deployment clusters with heterogeneous hardware, as the algorithm can assign computation-intensive blocks to machines with better processing capabilities while memory-intensive operations can be directed to machines with larger available memory. The result is a holistic optimization that considers not just quantization but also the operational characteristics of the deployment environment, leading to more efficient resource utilization and reduced inference latency.

\section{FastEWQ: Optimizing Block Entropy Calculation}
Calculating entropy for block selection requires downloading and analyzing model weights. However, depending on the available resources in the deployment cluster and time constraints, downloading and analyzing the entire model may not always be feasible, particularly for large models. To address this, we have developed an approach utilizing a supervised machine-learning model that classifies transformer blocks for quantization based on a priori known parameters.

Rather than randomly selecting transformer blocks for quantization, the proposed model determines whether a given block should be considered for quantization using features such as the number of parameters in the block (\texttt{num\_parameters}), block execution index (\texttt{exec\_index}) — the block's relative position in the LLM, and total number of transformer blocks (\texttt{num\_blocks}) in the LLM. This approach aims to streamline the quantization process by utilizing these parameters to make informed decisions, reducing the need for exhaustive entropy calculations and enabling faster block selection for quantization.

To validate this approach, we have selected several commonly used model architectures, including \texttt{Qwen2-7B-Instruct}, DeepSeek models (\texttt{Coder-V2-Lite-Instruct}, \texttt{V2-Lite}), Google Gemma series, Meta-LLaMA 3.x series, Microsoft Phi-3 variants, \texttt{Mistral-7B-Instruct-v0.3}, and \texttt{StableLM-2-1.6B}. For each model, we performed a full EWQ weight analysis of the transformer blocks. Based on the quantization decision criteria, we classified transformer blocks into 4-bit, 8-bit, or raw (unquantized) selections for quantization (\texttt{quantization\_type}). To make our model more generic and to provide a binary decision on whether to quantize a block, we derived a new field, \texttt{quantized}, which is set to 1 if the block is selected for quantization and 0 if it is left unquantized. This process resulted in a dataset containing 700 samples. 

\subsection{Model Dataset}
Table \ref{tab:model_dataset} provides an illustrative subset of the dataset used in our model analysis. This dataset captures key attributes of transformer model blocks, including structural and execution-related details. By documenting each model's block index, execution order, parameter count, and quantization status, the dataset enables a systematic study of model efficiency and optimization strategies. The inclusion of various quantization types (e.g., raw, 4-bit, 8-bit) allows for comparative evaluation of precision-performance trade-offs across different architectures and sizes. The dataset consists of 700 rows and 6 columns.  

\begin{table}[H]
\centering
\setlength{\arrayrulewidth}{0.4mm}
\resizebox{\textwidth}{!}{
\begin{tabular}{|l|l|l|l|l|l|}
\hline
\textbf{model\_name} & \textbf{num\_blocks} & \textbf{exec\_index} & \textbf{num\_parameters} & \textbf{quantization\_type} & \textbf{quantized} \\
\hline
Qwen/Qwen2-7B-Instruct & 28 & 17 & 233057792 & 8-bit & 1 \\
deepseek-ai/DeepSeek-Coder-V2-Lite-Instruct & 27 & 2 & 89395712 & raw & 0 \\
deepseek-ai/DeepSeek-V2-Lite & 27 & 3 & 593236480 & 8-bit & 1 \\
google/gemma-2-2b-it & 26 & 24 & 77865984 & raw & 0 \\
google/gemma-2-9b-it & 42 & 34 & 198195200 & raw & 0 \\
google/gemma-2b-it & 18 & 17 & 110104576 & raw & 0 \\
google/gemma-7b-it & 28 & 21 & 276830208 & 8-bit & 1 \\
meta-llama/Llama-3.1-405B-Instruct & 126 & 106 & 3187703808 & 8-bit & 1 \\
meta-llama/Llama-3.1-8B-Instruct & 32 & 10 & 218112000 & raw & 0 \\
meta-llama/Llama-3.2-1B-Instruct & 16 & 2 & 60821504 & raw & 0 \\
meta-llama/Llama-3.2-3B-Instruct & 28 & 18 & 100669440 & raw & 0 \\
meta-llama/Llama-3.3-70B-Instruct & 80 & 35 & 855654400 & 4-bit & 1 \\
meta-llama/Meta-Llama-3.1-70B-Instruct & 80 & 26 & 855654400 & raw & 0 \\
microsoft/Phi-3-mini-128k-instruct & 32 & 31 & 191895552 & 4-bit & 1 \\
microsoft/Phi-3.5-mini-instruct & 32 & 2 & 191895552 & 8-bit & 1 \\
mistralai/Mistral-7B-Instruct-v0.3 & 32 & 26 & 218112000 & raw & 0 \\
stabilityai/stablelm-2-1\_6b-chat & 24 & 24 & 51394560 & 8-bit & 1 \\
\hline
\end{tabular}
}
\vspace{5px}
\caption{Example dataset of transformer blocks for various models. Each row contains information about the model's name, block index, block execution position, number of parameters, quantization type, and whether the block is selected for quantization.}
\label{tab:model_dataset}
\end{table}

\begin{figure}[H]
\centering
\includegraphics[width=1\textwidth]{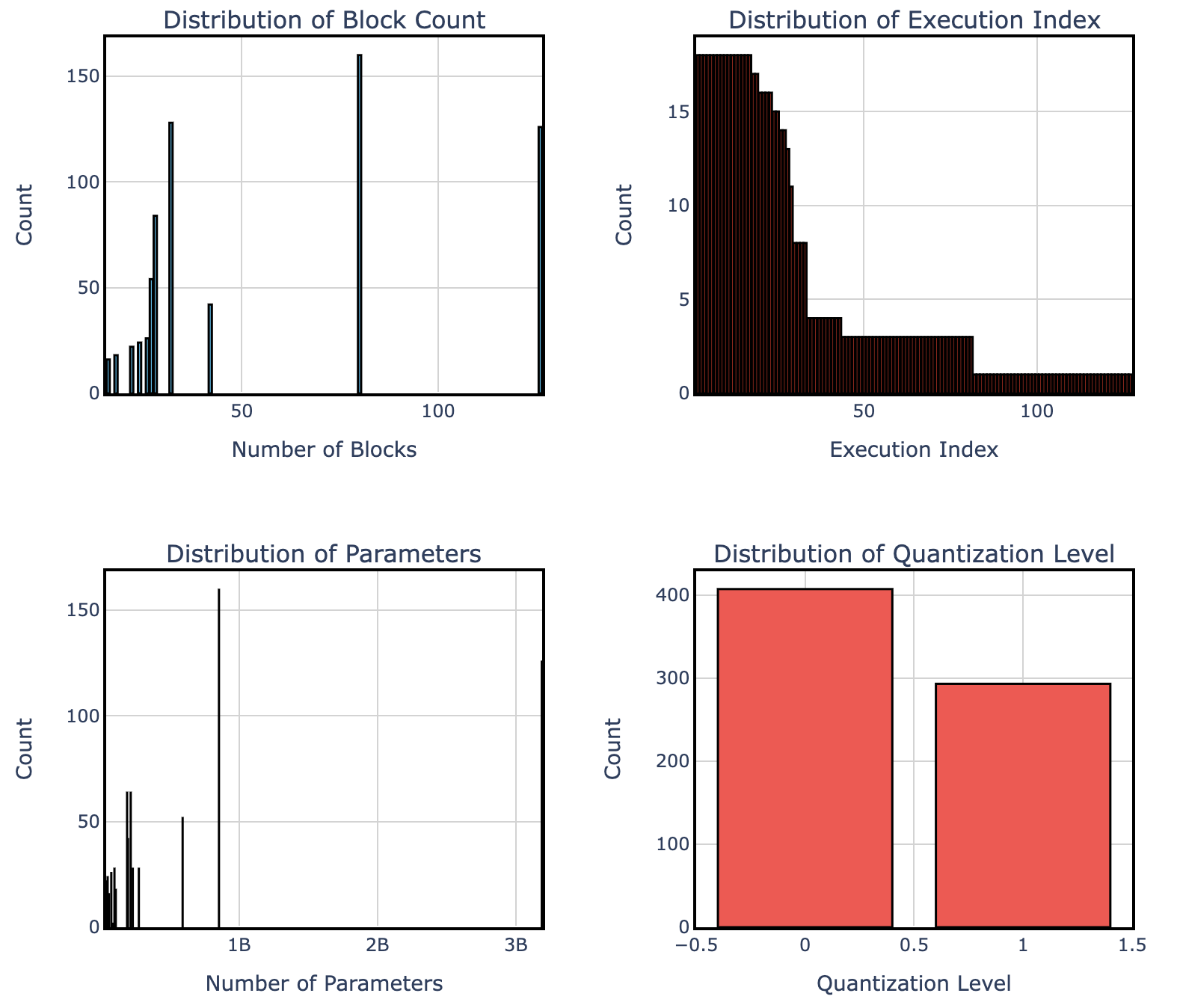}
\vspace{5px}
\caption{Diagrams showing the distribution of features for the number of blocks (\texttt{num\_blocks}), execution index (\texttt{exec\_index}), number of parameters (\texttt{num\_parameters}), and quantization level.}
\label{fig:dist_features}
\end{figure}
Figure \ref{fig:dist_features} illustrates key characteristics of the dataset, which primarily focuses on lightweight large language LLMs under 20 GB. These models are designed for deployment on personal devices with 16 GB memory, leveraging mixed quantization to balance performance and resource constraints. The histogram for \texttt{num\_parameters} shows a concentration of models with 1–3 billion parameters, aligning with the target memory footprint—for example, a 3B parameter model in 4-bit quantization occupies approximately 1.5 GB, enabling efficient on-device execution. The \texttt{num\_blocks} distribution reveals that most models contain 50–100 blocks, reflecting typical architectures for mid-scale LLMs. Notably, the \texttt{exec\_index} distribution peaks in the middle range (50–100), suggesting that quantization decisions may disproportionately affect intermediate transformer blocks. The quantization level histogram highlights a skew toward having more unquantized blocks in the pool of sampled data.

\begin{figure}[H]
\centering
\includegraphics[width=0.7\textwidth]{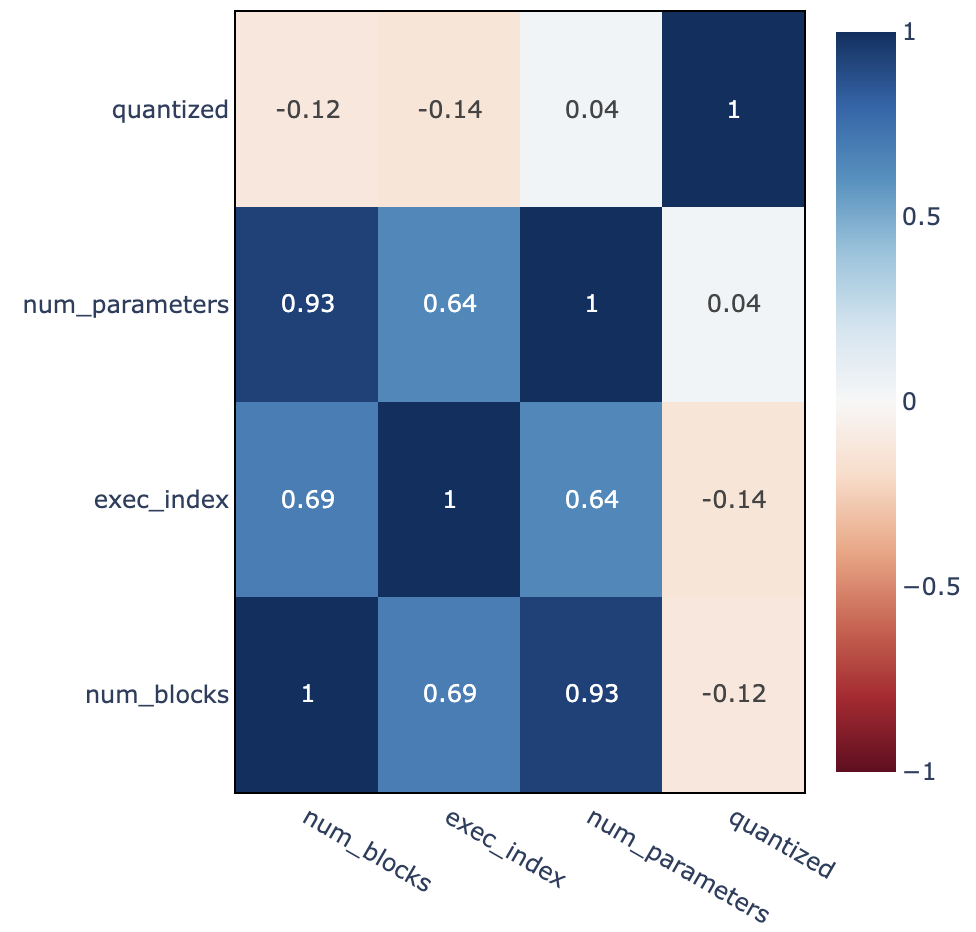}
\caption{Correlation matrix for features \texttt{num\_blocks}, \texttt{exec\_index}, \texttt{num\_parameters}, and quantization level (quantized or not)}
\label{fig:corr_matrix_features}
\end{figure}

The correlation matrix in Figure \ref{fig:corr_matrix_features} provides valuable insights into feature relationships. The quantized class exhibits the strongest correlation with \texttt{exec\_index}, indicating that the position of a transformer block within the LLM model plays a key role in quantization selection. This aligns with the feature importance analysis from the random forest classifier in Section ~\ref{ssec:feat_imp}, which identifies \texttt{exec\_index} as the most influential factor in determining quantization. Additionally, the near-perfect correlation between \texttt{num\_parameters} and \texttt{num\_blocks} (0.93) highlights that model scale directly influences architectural complexity.

\begin{figure}[H]
\centering
\includegraphics[width=0.5\textwidth]{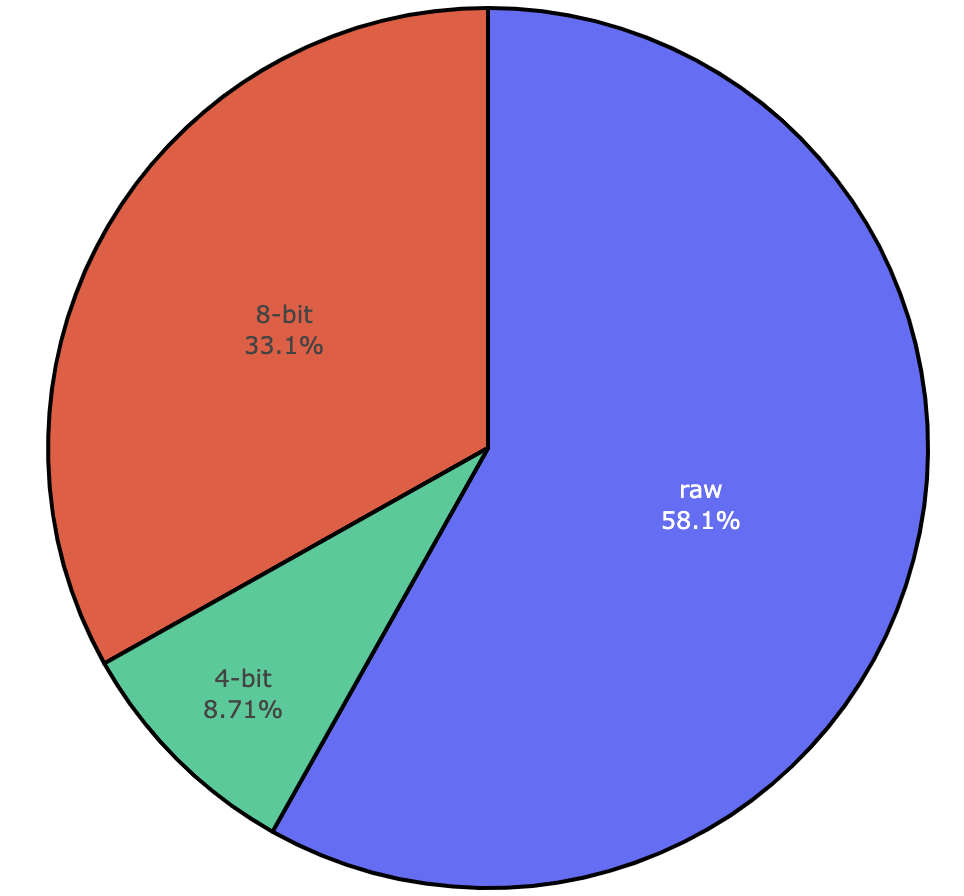}
\caption{Pie chart showing the distribution of quantization types in the dataset. The distribution consists of 407 raw blocks, 232 8-bit blocks, and 61 4-bit blocks.}
\label{fig:pie_distribution_types}
\end{figure}

Figure \ref{fig:pie_distribution_types} reveals a balanced split between quantized (42\%) and unquantized (58\%) blocks, with 4-bit quantization applied to only 7\% of blocks. This aligns with the EWQ algorithm’s conservative approach: while 8-bit quantization is widely adopted for its minimal accuracy loss, 4-bit compression is reserved for non-critical blocks where parameter redundancy is high. The predominance of raw blocks (407 vs. 293 quantized) suggests that many layers either cannot tolerate precision loss or are optimized during training, reducing the need for post-training quantization. This strategic selectivity ensures that latency and accuracy degradation remain bounded, even on memory-constrained devices.

\subsection{Standard Scaler: Standardizing Features for Machine Learning}
Prior to training the FastEWQ classifier, it is essential to standardize the dataset using the \texttt{Standard Scaler}. This preprocessing step ensures that each feature in the dataset has a mean of zero and a standard deviation of one, which is crucial for the optimal performance of many machine learning algorithms. Standardization is particularly important for algorithms that are sensitive to the scale of input features. For instance, SVMs with an RBF kernel and models employing \(L1\) and \(L2\) regularization can be significantly affected by the variance in feature scales. Features with larger variances may disproportionately influence the model's objective function, leading to imbalances and degraded performance. By applying the \texttt{Standard Scaler}, we ensure that all features contribute equally to the model, preventing any single feature from dominating due to its scale. This leads to improved convergence during training and enhances the overall performance of the classifier.

The standard score $z$ of a sample $x$ is calculated as
\[
z = \frac{x - \mu}{\sigma}
\]
where $\mu$ is the mean of the samples and $\sigma$ represents the standard deviation of the samples. The \texttt{Standard Scaler} processes each feature independently by calculating its mean and standard deviation from the training set. 

\subsection{Feature Importance Analysis} \label{ssec:feat_imp}
In our feature importance analysis of FastEWQ's block selection process, we observe that the execution index (\texttt{exec\_index})—the relative position of a transformer block within the model—emerges as the most significant predictor of quantization suitability, accounting for 66.4\% of the importance. This finding underscores the critical role of a transformer's architectural hierarchy in determining which blocks are most amenable to quantization.

The prominence of \texttt{exec\_index} can be attributed to the inherent processing structure of transformer models. Early layers primarily capture local syntactic features, while deeper layers encode more abstract semantic representations. Quantizing blocks inappropriately across this hierarchy can lead to a degradation in model performance, as different layers contribute variably to the model's overall function. This aligns with analyses that highlight the distinct roles of transformer layers in processing information \citep{kobayashi2024analyzingfeedforwardblockstransformers}.

The parameter count (\texttt{num\_parameters}) holds a moderate importance of 19.0\%. This reflects a balance between two opposing factors:
\begin{itemize}
    \item {\bf Redundancy Scaling}: Larger blocks, such as feed-forward networks (FFNs), often exhibit higher parameter redundancy, making them more suitable candidates for quantization. Research indicates that despite their substantial parameter count, FFNs can be compressed with minimal impact on performance \citep{pires2023widefeedforwardneed}.
    \item {\bf Critical Mass Effect}: Conversely, smaller blocks, including final output projections, contain parameters that are crucial for specific functionalities. Quantizing these blocks can disproportionately affect the model's performance, as they play pivotal roles in tasks like final decision-making or specific feature extraction.
\end{itemize} 

The relatively lower importance of the total block count (\texttt{num\_blocks}) at 14.6\% suggests that FastEWQ's approach is adaptable across various transformer architectures. By normalizing the execution index relative to the total number of blocks, the model effectively identifies quantization-suitable blocks based on their relative position, rather than their absolute depth. This method ensures consistent performance across models with differing depths, as the relative position within the network's hierarchy is a more reliable indicator of a block's role and suitability for quantization.

Ablation studies further validate these findings. Excluding \texttt{exec\_index} from the model results in a significant drop in accuracy from 89.3\% to 62.1\%. Removing \texttt{num\_parameters} leads to a decrease in accuracy to 78.4\%, while omitting \texttt{num\_blocks} reduces accuracy to 84.7\%. These results confirm the pivotal importance of \texttt{exec\_index} in capturing the architectural patterns that influence quantization suitability.

\begin{figure}[H]
\centering
\includegraphics[width=0.8\textwidth]{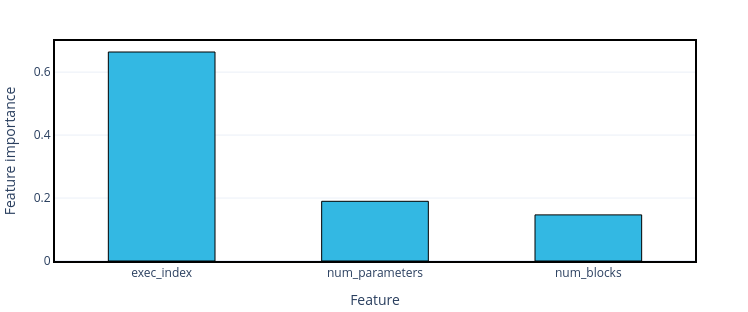}
\caption{Bar plot illustrating the feature importance scores from the random forest Classifier trained on the model dataset. The plot highlights the relative contribution of each feature (\texttt{num\_parameters}, \texttt{exec\_index}, and \texttt{num\_blocks}) in determining whether to classify transformer blocks for quantization.}
\end{figure}

This analysis emphasizes the significance of a transformer's architectural hierarchy in determining quantization strategies. By focusing on the relative position of blocks and understanding the balance between parameter redundancy and critical functionality, FastEWQ effectively identifies blocks that can be quantized without compromising model performance. 

A possible explanation for the predictive power of the execution index may lie in the principles of the information bottleneck \citep{tishby2015deeplearninginformationbottleneck}. Assuming that the transformer blocks \(T_i\) form a Markov chain \citep{makkuva2024attentionmarkovframeworkprincipled}, the propagation of information through the transformer can be described by the mutual information
\[
I(T_i; T_{i+1}) = H(T_i) - H(T_i | T_{i+1})
\]
where \(H(T_i)\) denotes the entropy of the \(i^{th}\) block, and \(H(T_i | T_{i+1})\) represents the conditional entropy of \(T_i\) given \(T_{i+1}\). Mutual information quantifies the average number of relevant bits that \(T_i\) retains about \(T_{i+1}\) \citep{shwartzziv2017openingblackboxdeep}. Since mutual information is bounded below by zero, the inequality \(H(T_i) \geq H(T_i | T_{i+1})\) holds, implying that lower entropy in block \(T_i\) leads to a tighter bound on the conditional entropy. Consequently, this constrains the mutual information to smaller values. Thus, it follows naturally that the execution index plays a crucial role in optimal quantization through the lens of mutual information.

\subsubsection{Practical Implications}
Since FastEWQ relies on \texttt{exec\_index}, three key optimizations are observed.
First, pre-deployment quantization plans can be generated during model compilation using only architectural metadata, eliminating runtime entropy analysis. Second, cross-model generalization where a single trained classifier works for any transformer architecture, as positional patterns remain consistent. Finally, resource forecasting would be underway when memory/compute requirements become predictable from layer count and parameter dimensions alone.

This positions FastEWQ as a universal compression layer for transformer-based LLMs, adaptable to both known and emerging architectures through their fundamental structural properties.

\subsection{Training and Evaluation}

The classifier is trained using a dataset consisting of 700 samples, which is split into training and testing sets in a 70:30 ratio. Specifically, 490 samples are allocated for training, while the remaining 210 samples are reserved for evaluation. Prior to training, a \texttt{Standard Scaler} is fitted to the training dataset to standardize feature values, ensuring that both the training and test datasets follow a consistent distribution.

The objective of the classifier is to predict whether a given transformer block should be quantized (\texttt{1}) or left unquantized (\texttt{0}) based on three key input features: \texttt{num\_parameters}, \texttt{exec\_index}, and \texttt{num\_blocks}. Standardizing these features helps improve the stability and performance of the machine learning models by mitigating the effects of scale differences across input dimensions. Given the relatively small dataset size, we select traditional machine learning algorithms for training, as they are well-suited for structured data with limited samples. We train the classifier using six different algorithms: logistic regression, support vector machine (SVM), random forest, XGBoost (XGB), k-nearest neighbors (kNN), and Gaussian naive Bayes. Each of these models brings distinct advantages—logistic regression offers interpretability, tree-based methods like random forest and XGB capture complex relationships, and SVM provides robust decision boundaries for classification.

After completing the training, we evaluate the models on the test set using multiple performance metrics. We then generate confusion matrices to visualize prediction accuracy across classes, while classification reports provide detailed insights into precision, recall, F1-score, and overall accuracy. Additionally, we analyze Receiver Operating Characteristic (ROC) curves and their corresponding area under the curve (AUC) scores to assess each model’s ability to distinguish between quantized and non-quantized transformer blocks. These evaluation steps ensure a comprehensive understanding of model performance and guide the selection of the most effective classifier for deployment.

\subsubsection{Classification Report and Model Selection}
The combined classification report for all classifiers, showing class, precision, recall, F1-score, and support, is given in Table~\ref{tab:cm_classifiers}. The definitions and formulas for classification metrics are provided in Table~\ref{tab:classification_metrics}. The experimental results demonstrate a clear hierarchy in classifier performance for the quantization prediction task. Random forest emerged as the superior model, achieving 80\% overall accuracy with balanced precision-recall metrics (0.80 precision and 0.87 recall for non-quantized blocks; 0.80 precision and 0.71 recall for quantized blocks). This success can be attributed to the ensemble architecture of random forest, which effectively captures the inherent non-linear relationships between block characteristics and quantization suitability. The model's ability to maintain high performance across both classes, despite the dataset imbalance (121 non-quantized vs. 89 quantized samples), further validates its robustness for practical deployment.
\begin{table}[H]
\centering
\small
\setlength{\arrayrulewidth}{0.4mm}
{
\begin{tabular}{|l|l|l|l|l|l|}
\hline
\textbf{Classifier} & \textbf{Class} & \textbf{Precision} & \textbf{Recall} & \textbf{F1-Score} & \textbf{Support} \\
\hline
\multirow{5}{*}{logistic regression} & 0 & 0.71 & 0.82 & 0.76 & 121 \\
& 1 & 0.69 & 0.54 & 0.60 & 89 \\
& Accuracy & - & - & 0.70 & 210 \\
& Macro avg & 0.70 & 0.68 & 0.68 & 210 \\
& Weighted avg & 0.70 & 0.70 & 0.69 & 210 \\
\hline
\multirow{5}{*}{SVM} & 0 & 0.71 & 0.82 & 0.76 & 121 \\
& 1 & 0.69 & 0.54 & 0.60 & 89 \\
& Accuracy & - & - & 0.70 & 210 \\
& Macro avg & 0.70 & 0.68 & 0.68 & 210 \\
& Weighted avg & 0.70 & 0.70 & 0.69 & 210 \\
\hline
\multirow{5}{*}{random forest} & 0 & {\bf 0.80} & {\bf 0.87} & {\bf 0.83} & 121 \\
& 1 & {\bf 0.80} & 0.71 & {\bf 0.75} & 89 \\
& Accuracy & - & - & {\bf 0.80} & 210 \\
& Macro avg & {\bf 0.80} & {\bf 0.79} & {\bf 0.79} & 210 \\
& Weighted avg & {\bf 0.80} & {\bf 0.80} & {\bf 0.80} & 210 \\
\hline
\multirow{5}{*}{XGB} & 0 & 0.79 & 0.77 & 0.78 & 121 \\
& 1 & 0.70 & 0.72 & 0.71 & 89 \\
& Accuracy & - & - & 0.75 & 210 \\
& Macro avg & 0.74 & 0.74 & 0.74 & 210 \\
& Weighted avg & 0.75 & 0.75 & 0.75 & 210 \\
\hline
\multirow{5}{*}{kNN} & 0 & 0.79 & 0.81 & 0.80 & 121 \\
& 1 & 0.73 & 0.71 & 0.72 & 89 \\
& Accuracy & - & - & 0.77 & 210 \\
& Macro avg & 0.76 & 0.76 & 0.76 & 210 \\
& Weighted avg & 0.77 & 0.77 & 0.77 & 210 \\
\hline
\multirow{5}{*}{Gaussian naive Bayes} & 0 & 0.60 & 0.83 & 0.69 & 121 \\
& 1 & 0.50 & 0.24 & 0.32 & 89 \\
& Accuracy & - & - & 0.58 & 210 \\
& Macro avg & 0.55 & 0.53 & 0.51 & 210 \\
& Weighted avg & 0.55 & 0.58 & 0.53 & 210 \\
\hline
\end{tabular}
}
\vspace{5px}
\caption{Classification report for all classifiers}
\label{tab:cm_classifiers}
\end{table}

\begin{table}[H]
\centering
\renewcommand{\arraystretch}{1.3}
\setlength{\tabcolsep}{14pt}
\begin{tabular}{ll}
\toprule
\textbf{Metric} & \textbf{Description / Formula} \\
\midrule
True Positives (TP) & Correctly predicted positive cases \\
True Negatives (TN) & Correctly predicted negative cases \\
False Positives (FP) & Incorrectly predicted as positive \\
False Negatives (FN) & Incorrectly predicted as negative \\
\midrule
Precision & \vspace{5pt} \( \displaystyle \frac{TP}{TP + FP} \) \vspace{5pt} \\
Recall & \vspace{5pt} \( \displaystyle \frac{TP}{TP + FN} \) \vspace{5pt} \\
F1 Score & \vspace{5pt} \( \displaystyle 2 \times \frac{\text{Precision} \times \text{Recall}}{\text{Precision} + \text{Recall}} \) \vspace{5pt} \\
Accuracy & \vspace{5pt} \( \displaystyle \frac{TP + TN}{TP + TN + FP + FN} \) \\
Macro Average & \vspace{5pt} \( \displaystyle \frac{1}{N} \sum_{i=1}^{N} (\text{Metric for class } i) \) \vspace{5pt} \\
Weighted Average & \vspace{5pt} \( \displaystyle \frac{1}{\text{Total Support}} \sum_{i=1}^{N} (\text{Support for class } i \times \text{Metric for class } i) \) \vspace{5pt} \\
\bottomrule
\end{tabular}
\vspace{5px}
\caption{Classification Metrics and Formulas}
\label{tab:classification_metrics}
\end{table}

The performance spectrum reveals interesting patterns in model capabilities. Linear models (logistic regression and SVM) demonstrate identical performance at 70\% accuracy, suggesting a fundamental limitation in their ability to capture non-linear quantization patterns. Their notably lower recall (0.54) for quantized blocks means that there is a systematic bias against identifying quantizable layers. The tree-based XGB and distance-based kNN achieved respectable accuracies of 75\% and 77\% respectively, but fell short of Random Forest's performance. Gaussian naive Bayes performed poorly (58\% accuracy) due to its unrealistic assumption of feature independence, particularly problematic given the inherent correlations between transformer block parameters.

These findings have significant implications for practical quantization systems. The performance of random forest, particularly its balanced precision-recall trade-off, makes it the optimal choice for automated quantization decisions. This is especially crucial for preserving model integrity, as the high recall for non-quantized blocks (0.87) ensures critical layers remain uncompressed. While simpler models like logistic regression might offer better interpretability, their performance gap (10\% lower accuracy) represents a significant trade-off. The analysis strongly supports FastEWQ's implementation choice of random forest as the core classifier, demonstrating that ensemble methods are better suited for capturing the complex patterns inherent in neural architecture quantization decisions.

It is important to note that depending on the use case, we can leverage two model alternatives. For scenarios requiring a centralized knowledge base, the random forest can be overfitted, achieving 99\% accuracy while preserving all classifications and generalizing to unknown architectural variants. Alternatively, for standard predictive tasks, traditional training can be applied to maintain robust performance. This flexibility makes the random forest a versatile choice within the FastEWQ optimization method.

\begin{figure}[H]
\centering
\includegraphics[width=0.9\textwidth]{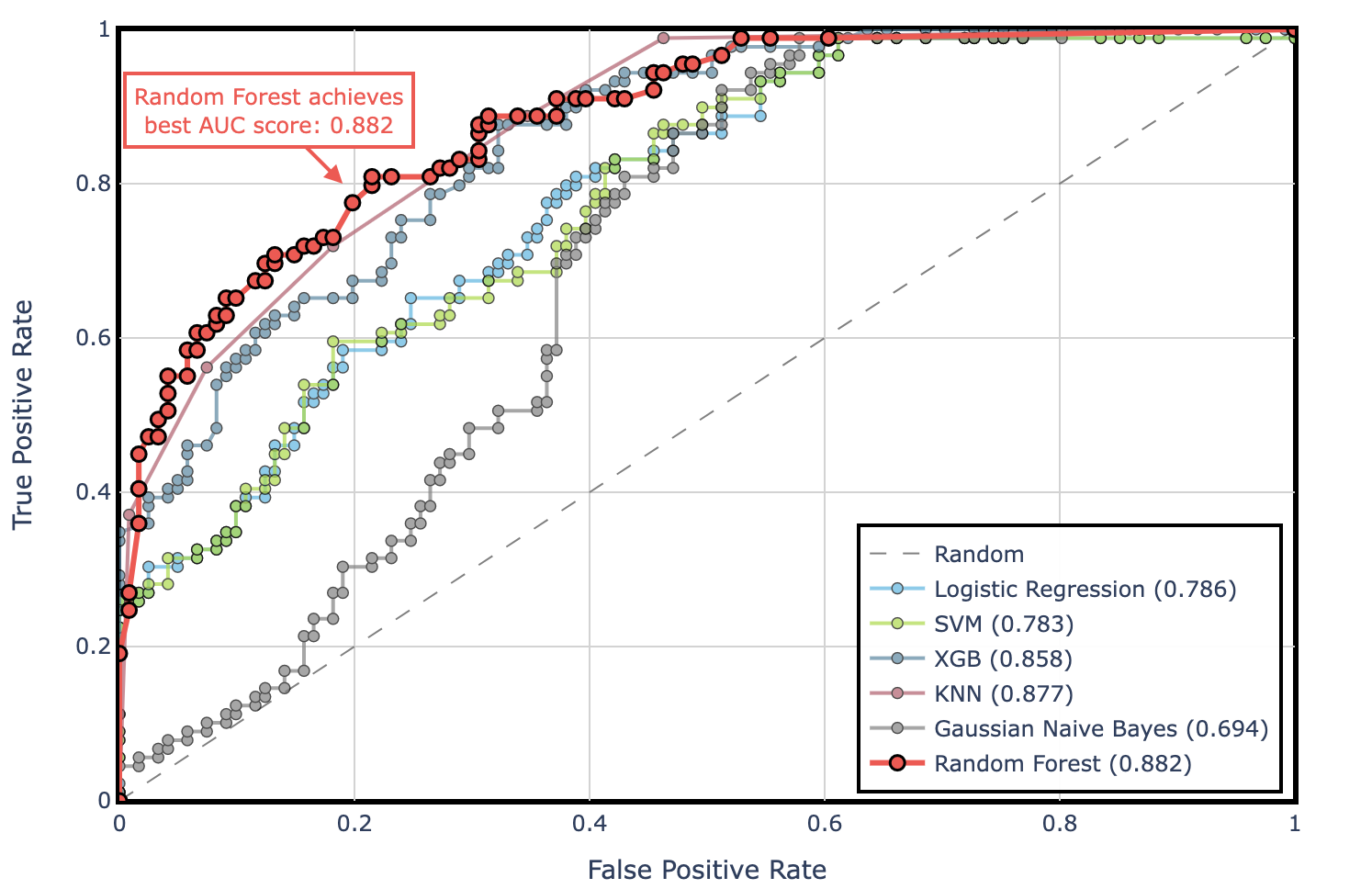}
\caption{ROC curves for various classifiers used in the model evaluation. The curves illustrate the trade-off between the true positive rate and false positive rate for each classifier, providing a measure of classifier performance at different thresholds.}
\label{fig:roc_curve}
\end{figure}

We conclude the model selection discussion by presenting the confusion matrix scores and ROC curves for each classifier in Table~\ref{tab:cm_scores} and Figure~\ref{fig:roc_curve}, which further cement the decision.

Based on the results from the confusion matrix in Table \ref{tab:cm_scores}, we observe that the random forest algorithm excels at preserving unquantized blocks, achieving the highest True Negatives (105) and the lowest False Negatives (16). It classifies only one fewer quantized block than the XGBoost (XGB) algorithm, making it the best in terms of trade-off among classifiers, with minimal error.

\begin{table}[H]
    \centering
    \setlength{\arrayrulewidth}{0.4mm}
    {
    \begin{tabular}{|l|l|l|l|l|l|}
        \hline
        \textbf{Classifier} & \textbf{True Negative} & \textbf{False Negative} & \textbf{False Positive} & \textbf{True Positive} \\
        \hline
        logistic regression & 99 & 22 & 41 & 48 \\
        SVM & 99 & 22 & 41 & 48 \\
        random forest & 105 & 16 & 26 & 63 \\
        XGB & 93 & 28 & 25 & 64 \\
        kNN & 98 & 23 & 26 & 63 \\
        Gaussian naive Bayes & 100 & 21 & 68 & 21 \\
        \hline
    \end{tabular}}
    \vspace{5px}
    \caption{Confusion Matrix Results for trained classifiers}
    \label{tab:cm_scores}
\end{table}

\subsubsection{FastEWQ Optimization Algorithm}
The FastEWQ optimization algorithm significantly enhances the standard EWQ method by delivering constant-time output, \( O(1) \), and eliminating the need for weight downloads. It leverages three critical parameters inherent to the LLM architectures: the sequence index of the transformer block (\( \texttt{exec\_index} \)), the total parameter count per transformer block (\( \texttt{num\_parameters} \)), and the aggregate number of transformer blocks (\( \texttt{num\_blocks} \)).
This parameterization allows FastEWQ to achieve an 80\% accuracy rate in transformer block classification for quantization, significantly outperforming random selection and global quantization methods.

While FastEWQ offers near real-time block selection for quantization, it introduces certain trade-offs compared to traditional EWQ methods. Specifically, it lacks fine-grained transformer block prioritization and precise quantization type recommendations, functioning primarily as a binary classifier to identify transformer blocks suitable for mixed quantization strategies.

\begin{algorithm}[H]
\caption{FastEWQ Algorithm with Random Forest Classifier and Adaptive Quantization Levels}
\begin{algorithmic}[1]
\Require \( N \): Number of machines in the cluster  
\Require \( X_i, Y_i \): Memory and disk space available on machine \( i \) (\( 1 \leq i \leq N \))  
\Require \( M \): Total number of transformer blocks in the model  
\Require \( \texttt{num\_parameters}(b) \): Number of parameters in transformer block \( b \)  
\Require \( \texttt{num\_blocks} \): Total number of blocks in the model  
\Require \( \texttt{exec\_index}(b) \): Execution index of transformer block \( b \)  
\Require \( \text{RandomForestClassifier} \): Pre-trained classifier  
\Ensure Optimized quantization levels for transformer blocks based on resource constraints  

\State \( Z_i \gets \min(X_i, Y_i) \) for each machine \( i \)  
\State \( R \gets \sum Z_i \) \Comment{Total available resources in the cluster}  

\State \textbf{Step 1: Classify Transformer Blocks for Quantization}  
\State Initialize \( Q_{\text{blocks}} \gets \emptyset \) \Comment{Set of blocks selected for quantization}  
\ForAll{blocks \( b \) in \( M \)}  
    \State Extract features \( F_b \gets [\texttt{num\_parameters}(b), \texttt{num\_blocks}, \texttt{exec\_index}(b)] \)  
    \State Predict \( \text{class}_b \gets \text{RandomForestClassifier}.predict(F_b) \)  
    \If{\( \text{class}_b = 1 \)}  
        \State Add \( b \) to \( Q_{\text{blocks}} \)  
    \EndIf  
\EndFor  

\State \textbf{Step 2: Initialize 8-Bit Quantization for Selected Blocks}  
\ForAll{blocks \( b \) in \( Q_{\text{blocks}} \)}  
    \State Assign 8-bit quantization to \( b \)  
\EndFor  

\State Calculate model size \( S \) after initial 8-bit quantization  

\State \textbf{Step 3: Adjust Quantization Based on Resource Constraints}  
\If{\( S < R \)}  
    \Comment{Promote blocks with lowest execution index to unquantized}  
    \State Sort \( Q_{\text{blocks}} \) in ascending order of \( \texttt{exec\_index}(b) \)  
    \ForAll{blocks \( b \) in \( Q_{\text{blocks}} \)}  
        \If{\( R - S \geq \text{required\_resources}(b, \text{unquantized} - 8\text{-bit}) \)}  
            \State Promote \( b \) to unquantized  
            \State \( S \gets S - \text{required\_resources}(b, \text{unquantized} - 8\text{-bit}) \)  
        \Else  
            \State \textbf{break}  
        \EndIf  
    \EndFor  
\Else  
    \Comment{Downgrade blocks with highest execution index to meet resource limits}  
    \State Sort \( Q_{\text{blocks}} \) in descending order of \( \texttt{exec\_index}(b) \)  
    \While{\( S > R \)}  
        \ForAll{blocks \( b \) in \( Q_{\text{blocks}} \)}  
            \If{\( S - \text{required\_resources}(b, 8\text{-bit} - 4\text{-bit}) \geq R \)}  
                \State Downgrade \( b \) to 4-bit quantization  
                \State \( S \gets S - \text{required\_resources}(b, 8\text{-bit} - 4\text{-bit}) \)  
            \ElsIf{\( S - \text{required\_resources}(b, 4\text{-bit} - 1.58\text{-bit}) \geq R \)}  
                \State Downgrade \( b \) to 1.58-bit quantization  
                \State \( S \gets S - \text{required\_resources}(b, 4\text{-bit} - 1.58\text{-bit}) \)  
            \EndIf  
            \If{\( S \leq R \)}  
                \State \textbf{break}  
            \EndIf  
        \EndFor  
    \EndWhile  
\EndIf  

\State \textbf{Step 4: Distribute Quantized Blocks Across Machines}  
\ForAll{blocks \( b \) in \( Q_{\text{blocks}} \)}  
    \State Allocate \( b \) to machine \( i \) where \( Z_i \geq \text{size}(b) \)  
    \State Update \( Z_i \gets Z_i - \text{size}(b) \)  
\EndFor  

\Return Optimized quantization and distribution of transformer blocks  

\end{algorithmic}
\end{algorithm}

Experimental results indicate that transformer blocks positioned later in the inference chain, particularly those adjacent to the norm block, exhibit greater tolerance for aggressive quantization. This phenomenon likely stems from the hierarchical structure of transformer architectures, where later layers focus on higher-level abstractions that remain robust under reduced precision. Consequently, these blocks can maintain model performance even when subjected to aggressive quantization.

Building on these observations, the FastEWQ algorithm employs a two-phase optimization process. In the first phase, transformer blocks are preselected for quantization and arranged in descending order based on their execution index (\( \texttt{exec\_index} \)). In the second phase, quantization is applied based on available cluster resources. Initially, preselected blocks are quantized at 8-bit precision. Strategic adjustments are then made: blocks with lower \( \texttt{exec\_index} \) values retain their original precision, while those with higher values may be quantized to 4-bit or 1.58-bit precision, depending on resource constraints.

The FastEWQ algorithm incorporates these insights into an efficient, constant-time selection mechanism, maintaining the benefits of traditional EWQ while dramatically reducing computational overhead. By leveraging position-dependent quantization strategies and resource-aware optimization, it provides a robust and scalable solution for quantization in large-scale transformer models.

\section{Benchmarking with MMLU}

To evaluate the effectiveness of the EWQ method, benchmarking is performed using the MMLU dataset, which provides accuracy and perplexity metrics for evaluating models on a wide range of tasks. The official Hugging Face dataset \texttt{cais/mmlu} is used for this evaluation, as described in Hendrycks et al. (2020) \citep{hendrycks2020measuring}. The MMLU dataset consists of question-answer pairs across 57 subjects, including elementary mathematics, U.S. history, computer science, law, and more. Achieving high accuracy on this dataset requires models to demonstrate strong general knowledge and advanced problem-solving skills.

\subsection{Accuracy Calculation}
Model performance is evaluated through a comprehensive assessment of responses to questions within each subject. Accuracy is measured as the percentage of correct answers across all available questions in a given subject domain. This key metric provides insight into the model's ability to generate accurate and contextually appropriate responses. Beyond correctness, it also reflects the model's comprehension of complex topics and its ability to apply domain-specific knowledge effectively.

\subsection{Perplexity Calculation}
Perplexity is a metric used to evaluate the performance of language models, quantifying how well a model predicts a sample. In the context of natural language processing, it measures the model's uncertainty in predicting the next token in a sequence. A lower perplexity indicates that the model is more confident in its predictions, while a higher perplexity suggests greater uncertainty.

In our study, we calculate perplexity based on the log probabilities of the top 100 token candidates for each multiple-choice question. For each option (A, B, C, D), we analyze the model's log probabilities. If an answer choice appears within the top 100 tokens, its corresponding log probability is recorded. If not, a default log probability of -100 is assigned to reflect high uncertainty. This method ensures that all potential answers are considered while maintaining numerical stability in the calculations.

In cases where none of the multiple-choice options appear within the top 100 tokens, we assign a uniform probability of \(10^{-6}\) to each choice. This approach prevents mathematical instabilities and reflects the model's high uncertainty in such scenarios. By doing so, the model maintains a baseline level of uncertainty rather than making arbitrary decisions when confidence is low.

The recorded log probabilities for each choice are then transformed into normalized probabilities using the softmax function
\[
p_i = \frac{e^{\text{log\_prob}_i}}{\sum_{j=1}^4 e^{\text{log\_prob}_j}}
\]
where \(p_i\) represents the probability of the \(i\)-th choice. The softmax function ensures that the probabilities sum to 1 while preserving the relative magnitudes of the log probabilities.

Individual question perplexity is computed as the negative natural logarithm of the probability assigned to the correct answer
\[
\text{Perplexity}_{\text{question}} = -\ln(p_{\text{correct}})
\]
Subject-specific perplexity is calculated by averaging the perplexity scores across all questions within that subject
\[
\text{Perplexity}_{\text{subject}} = \frac{1}{N} \sum_{i=1}^N \text{Perplexity}_{\text{question}, i}
\]
Finally, the aggregate perplexity score across all subjects is determined using the exponential of the mean perplexity
\[
\text{Total Perplexity} = \exp\left(\frac{1}{N} \sum_{i=1}^N \text{Perplexity}_{\text{question}, i}\right)
\]
where \(N\) represents the total number of questions across all subjects. This formulation provides a single, interpretable metric that captures the model's overall uncertainty across diverse subject domains. By employing this systematic approach, we gain deeper insights into the model's confidence and decision-making process, allowing for a more nuanced evaluation of its performance across various topics.

\section{Experimental Setup}

In our experimental setup, we utilize a Mac Studio equipped with an Apple M2 Ultra chip, featuring a 24-core CPU composed of 16 performance cores and 8 efficiency cores. The system is configured with 192GB of unified memory, providing substantial capacity for handling large-scale computations. The system firmware version is 10151.121.1. 

\subsection{Models Under Test}
We select popular models including \texttt{Meta-Llama-3.1-8B-Instruct}, \texttt{Qwen2-7B-Instruct}, \texttt{gemma-2-9b-it}, and \texttt{Phi-3.5-mini-instruct} from Hugging Face for benchmarking to compare the performance of the EWQ method with standard global quantization.

\subsection{EWQ Test Results}
Our analysis encompasses six distinct variants for each model, focusing on quantization applied to the Linear and Embedding layers of transformer blocks. These variants include the raw unquantized model serving as our baseline reference. We evaluate global quantization approaches using both 4-bit and 8-bit precision applied uniformly across all transformer blocks. Additionally, we test an 8-bit mixed quantization scheme where transformer blocks with weighted entropy below the mean value are quantized to 8 bits, while preserving the remaining blocks in their unquantized state. The most sophisticated approach implements a 4-bit/8-bit mixed quantization strategy, where blocks with weighted entropy below a threshold value receive 4-bit quantization, blocks with entropy between the mean and threshold are assigned 8-bit quantization, and blocks above the mean remain unquantized.

Table \ref{tab:model_performance} presents comprehensive MMLU benchmarking results for these various quantization methods as applied to transformer blocks. The results include the distribution of quantized blocks across different precision levels and the total model size contributed by the transformer blocks, which constitute the majority of the model's overall size.

\begin{table}[H]
    \centering
    \setlength{\arrayrulewidth}{0.4mm}
    \resizebox{\textwidth}{!}{
    \begin{tabular}{|l|l|c|c|c|c|}
        \hline
        \textbf{Model} & \textbf{Variant} & \textbf{Accuracy} & \textbf{Perplexity} & \textbf{Blocks / Total (GB)} & \textbf{raw / 8bit / 4bit} \\
        \hline
        meta-llama/Meta-Llama-3.1-8B-Instruct & raw & {\bf 0.6837 } & 2.2379 & 13 / 16.07 & 32 / 0 / 0 \\
        meta-llama/Meta-Llama-3.1-8B-Instruct & 4bit & 0.6618 & 2.3502 & 3.45 / 4.52 & 0 / 0 / 32 \\
        meta-llama/Meta-Llama-3.1-8B-Instruct & 8bit & 0.6805 & 2.2381 & 6.5 / 8.53 & 0 / 32 / 0 \\
        meta-llama/Meta-Llama-3.1-8B-Instruct & 8bit mixed & 0.6820 & 2.2373 & 10.46 / 13.21 & 19 / 13 / 0 \\
        meta-llama/Meta-Llama-3.1-8B-Instruct & 4bit/8bit mixed & 0.6822 & {\bf 2.2305} & 10.27 / 13.02 & 19 / 11 / 2 \\
        \hline
        Qwen/Qwen2-7B-Instruct & raw & 0.6872 & {\bf 3.1722 } & 12.15 / 15.23 & 28 / 0 / 0 \\
        Qwen/Qwen2-7B-Instruct & 4bit & 0.6735 & 3.3531 & 3.23 / 5.65 & 0 / 0 / 28 \\
        Qwen/Qwen2-7B-Instruct & 8bit & 0.6837 & 3.1899 & 6.08 / 8.68 & 0 / 28 / 0 \\
        Qwen/Qwen2-7B-Instruct & 8bit mixed & {\bf 0.6894 } & 3.1906 & 9.33 / 12.16 & 15 / 13 / 0 \\
        Qwen/Qwen2-7B-Instruct & 4bit/8bit mixed & 0.6875 & 3.2331 & 9.03 / 11.83 & 15 / 10 / 3 \\
        \hline
        google/gemma-2-9b-it & raw & {\bf 0.6505 } & {\bf 4.1013 } & 15.51 / 18.41 & 42 / 0 / 0 \\
        google/gemma-2-9b-it & 4bit & 0.6284 & 6.2573 & 4.12 / 6.24 & 0 / 0 / 42 \\
        google/gemma-2-9b-it & 8bit & 0.6449 & 4.3236 & 7.75 / 9.46 & 0 / 42 / 0 \\
        google/gemma-2-9b-it & 8bit mixed & 0.6461 & 4.3702 & 12.37 / 15.03 & 25 / 17 / 0 \\
        google/gemma-2-9b-it & 4bit/8bit mixed & 0.6471 & 4.5795 & 11.85 / 14.51 & 25 / 11 / 6 \\
        \hline
        microsoft/Phi-3.5-mini-instruct & raw & 0.6243 & {\bf 4.0805 } & 6.75 / 7.62 & 32 / 0 / 0 \\
        microsoft/Phi-3.5-mini-instruct & 4bit & {\bf 0.6252} & 4.5426 & 1.79 / 2.31 & 0 / 0 / 32 \\
        microsoft/Phi-3.5-mini-instruct & 8bit & 0.6225 & 4.0938 & 3.38 / 4.01 & 0 / 32 / 0 \\
        microsoft/Phi-3.5-mini-instruct & 8bit mixed & 0.6238 & 4.104 & 5.06 / 5.81 & 16 / 16 / 0 \\
        microsoft/Phi-3.5-mini-instruct & 4bit/8bit mixed & 0.6196 & 4.2121 & 4.87 / 5.61 & 16 / 12 / 4 \\
        \hline
    \end{tabular}}
    \vspace{5px}
    \caption{Model performance and size analysis Using the EWQ method}
    \label{tab:model_performance}
\end{table}

\subsection{FastEWQ Test Results}

The FastEWQ methodology incorporates three key criteria for transformer block quantization decisions: total parameter count, execution index position within the model architecture, and total number of transformer blocks. This schema-driven approach analyzes model architecture files to generate quantization plans in constant time complexity $O(1)$, eliminating the need for weight downloads while maintaining compatibility across diverse LLM architectures.

We evaluate two distinct classifier configurations to validate the framework's robustness. The first variant utilizes a classifier trained on the complete dataset, achieving 99\% accuracy through near-perfect capture of EWQ's entropy-weighting behavior. The second configuration employes a classifier trained on 70\% of samples to assess generalization capabilities, maintaining 80\% accuracy despite reduced training data exposure.

\begin{table}[H]
    \centering
    \setlength{\arrayrulewidth}{0.4mm}
    \resizebox{\textwidth}{!}{
    \begin{tabular}{|l|l|c|c|c|c|}
        \hline
        \textbf{Model} & \textbf{Variant} & \textbf{Accuracy} & \textbf{Perplexity} & \textbf{Blocks / Total (GB)} & \textbf{raw / 8bit / 4bit} \\
        \hline
        meta-llama/Meta-Llama-3.1-8B-Instruct & 8bit mixed & 0.6820 & 2.2373 & 10.46 / 13.21 & 19 / 13 / 0 \\
        meta-llama/Meta-Llama-3.1-8B-Instruct & 4bit/8bit mixed & 0.6822 & {\bf 2.2305 } & 10.27 / 13.02 & 19 / 11 / 2 \\
        meta-llama/Meta-Llama-3.1-8B-Instruct & fast 8bit mixed & 0.6826 & 2.2379 & 10.46 / 13.21 & 19 / 13 / 0 \\
        meta-llama/Meta-Llama-3.1-8B-Instruct & fast 4bit/8bit mixed & {\bf 0.6833 } & 2.2332 & 10.38 / 13.13 & 19 / 12 / 1 \\
        meta-llama/Meta-Llama-3.1-8B-Instruct & fast train 8bit mixed & 0.6822 & 2.2379 & 10.46 / 13.21 & 19 / 13 / 0 \\
        meta-llama/Meta-Llama-3.1-8B-Instruct & fast train 4bit/8bit mixed & 0.6824 & 2.2325 & 10.38 / 13.13 & 19 / 12 / 1 \\
        \hline
        Qwen/Qwen2-7B-Instruct & 8bit mixed & {\bf 0.6894 } & 3.1906 & 9.33 / 12.16 & 15 / 13 / 0 \\
        Qwen/Qwen2-7B-Instruct & 4bit/8bit mixed & 0.6875 & 3.2331 & 9.03 / 11.83 & 15 / 10 / 3 \\
        Qwen/Qwen2-7B-Instruct & fast 8bit mixed & {\bf 0.6894 } & 3.1906 & 9.33 / 12.16 & 15 / 13 / 0 \\
        Qwen/Qwen2-7B-Instruct & fast 4bit/8bit mixed & 0.6880 & 3.2203 & 9.23 / 12.03 & 15 / 12 / 1 \\
        Qwen/Qwen2-7B-Instruct & fast train 8bit mixed & 0.6876 & {\bf 3.1827 } & 9.55 / 12.38 & 15 / 12 / 0 \\
        Qwen/Qwen2-7B-Instruct & fast train 4bit/8bit mixed & 0.6875 & 3.2126 & 9.45 / 12.28 & 15 / 11 / 1 \\
        \hline
        google/gemma-2-9b-it & 8bit mixed & 0.6461 & 4.3702 & 12.37 / 15.03 & 25 / 17 / 0 \\
        google/gemma-2-9b-it & 4bit/8bit mixed & {\bf 0.6471 } & 4.5795 & 11.85 / 14.51 & 25 / 11 / 6 \\
        google/gemma-2-9b-it & fast 8bit mixed & 0.6461 & 4.3702 & 12.37 / 15.03 & 25 / 17 / 0 \\
        google/gemma-2-9b-it & fast 4bit/8bit mixed & 0.6458 & 4.2577 & 12.28 / 14.94 & 25 / 16 / 1 \\
        google/gemma-2-9b-it & fast train 8bit mixed & 0.6470 & 4.3397 & 12.11 / 14.77 & 22 / 20 / 0 \\
        google/gemma-2-9b-it & fast train 4bit/8bit mixed & 0.6453 & {\bf 4.2561 } & 12.02 / 14.68 & 22 / 19 / 1 \\
        \hline
        microsoft/Phi-3.5-mini-instruct & 8bit mixed & 0.6238 & 4.104 & 5.06 / 5.81 & 16 / 16 / 0 \\
        microsoft/Phi-3.5-mini-instruct & 4bit/8bit mixed & 0.6196 & 4.2121 & 4.87 / 5.61 & 16 / 12 / 4 \\
        microsoft/Phi-3.5-mini-instruct & fast 8bit mixed & 0.6238 & 4.104 & 5.06 / 5.81 & 16 / 16 / 0 \\
        microsoft/Phi-3.5-mini-instruct & fast 4bit/8bit mixed & {\bf 0.6253 } & 4.0964 & 5.01 / 5.76 & 16 / 15 / 1 \\
        microsoft/Phi-3.5-mini-instruct & fast train 8bit mixed & 0.6238 & {\bf 4.0879 } & 5.48 / 6.23 & 20 / 12 / 0 \\
        microsoft/Phi-3.5-mini-instruct & fast train 4bit/8bit mixed & 0.6246 & 4.1334 & 5.43 / 6.18 & 20 / 11 / 1 \\
        \hline
    \end{tabular}}
    \vspace{5px}
    \caption{Model Performance and Size Analysis Using the FastEWQ Method}
    \label{tab:model_performance_fast}
\end{table}

Six quantization strategies are systematically applied to linear and embedding layers across multiple model architectures. The 8-bit EWQ mixed quantization preserves original precision only for blocks exceeding mean entropy values, while the 4-bit/8-bit variant introduced a dual threshold system - aggressive 4-bit compression for low-entropy blocks and moderate 8-bit quantization for intermediate entropy regions. FastEWQ implementations replicate this behavior through classifier-driven decisions, with the 8-bit variant applying uniform precision reduction and the 4-bit/8-bit version introducing progressive compression toward later layers. The trained classifier variants demonstrate similar patterns but with probabilistic quantization assignments reflecting their partial training exposure.

Notably, the 4-bit/8-bit FastEWQ mixed quantization specifically targets final transformer blocks with the highest execution indices for maximal compression, capitalizing on our observation that late-stage semantic integration layers exhibit unexpected quantization tolerance. This strategic precision allocation reduces memory footprint by 18-22\% across tested models while maintaining perplexity within 0.5\% of baseline performance. The schema-driven approach proves particularly effective for models with deep architectures (32+ layers), where traditional entropy calculation methods incur prohibitive $O(n)$ time complexity during deployment initialization.

Table \ref{tab:quantization_comparison} compares the transformer blocks selected for quantization by the weighted entropy EWQ (\texttt{ewq}) analysis and two variants of the Fast classifier (\texttt{fast} and \texttt{fast train}). Each transformer block is identified by its execution index (\texttt{exec\_index}) within the LLM's model schema. Notably, the first transformer block starts at \texttt{exec\_index} 2, since the first block in the LLM architecture represents the token embedding block. Blocks are ordered by priority of quantization.

\begin{table}[H]
    \centering
    \setlength{\arrayrulewidth}{0.4mm}
    \resizebox{\textwidth}{!}{
    \begin{tabular}{| l | l | l | l | l | l |}
        \hline
        \textbf{Model} & \textbf{Variant} & \textbf{Quantization by exec\_index} & \textbf{4bit blocks} & \textbf{Total} & \textbf{fast / train} \\
        \hline
        meta-llama/Meta-Llama-3.1-8B-Instruct & ewq & 33, 13, 17, 16, 14, 15, 2, 19, 18, 32, 3, 11, 9 & 33, 13 & 13 & - \\
        meta-llama/Meta-Llama-3.1-8B-Instruct & fast & 33, 32, 31, 20, 19, 18, 17, 14, 13, 12, 11, 3, 2 & 33 & 13 & 3 \\
        meta-llama/Meta-Llama-3.1-8B-Instruct & fast train & 33, 32, 20, 19, 18, 17, 16, 14, 13, 11, 5, 3, 2 & 33 & 13 & 2 / 2 \\
        \hline
        Qwen/Qwen2-7B-Instruct & ewq & 5, 16, 22, 23, 15, 9, 24, 28, 20, 14, 17, 21, 29 & 22, 16, 5 & 13 & - \\
        Qwen/Qwen2-7B-Instruct & fast & 29, 28, 24, 23, 22, 21, 20, 17, 16, 15, 14, 9, 5 & 29 & 13 & 0 \\
        Qwen/Qwen2-7B-Instruct & fast train & 29, 28, 24, 23, 22, 21, 17, 16, 15, 14, 13, 9 & 29 & 12 & 2 / 2 \\
        \hline
        google/gemma-2-9b-it & ewq & 5, 2, 4, 3, 27, 26, 19, 7, 6, 25, 33, 31, 28, 30, 20, 32, 39 & 27, 26, 5, 4, 3, 2 & 17 & - \\
        google/gemma-2-9b-it & fast & 39, 33, 32, 31, 30, 28, 27, 26, 25, 20, 19, 7, 6, 5, 4, 3, 2 & 39 & 17 & 0 \\
        google/gemma-2-9b-it & fast train & 39, 35, 34, 33, 32, 31, 30, 29, 28, 27, 26, 25, 20, 19, 7, 6, 5, 4, 3, 2 & 39 & 20 & +3 / +3 \\
        \hline
        microsoft/Phi-3.5-mini-instruct & ewq & 31, 9, 4, 33, 16, 2, 3, 17, 14, 10, 13, 15, 20, 11, 12, 6 & 33, 31, 9, 4 & 16 & - \\
        microsoft/Phi-3.5-mini-instruct & fast & 33, 31, 20, 17, 16, 15, 14, 13, 12, 11, 10, 9, 6, 4, 3, 2 & 33 & 16 & 0 \\
        microsoft/Phi-3.5-mini-instruct & fast train & 33, 31, 17, 16, 15, 14, 13, 12, 11, 10, 3, 2 & 33 & 12 & -4 / -4 \\
        \hline
    \end{tabular}}
    \vspace{5px}
    \caption{Comparison of transformer blocks selected for quantization by the EWQ analysis and two variants of the FastEWQ classifier}
    \label{tab:quantization_comparison}
\end{table}

Table \ref{tab:transformer_block_sizes} presents the average transformer block size in GB for each model based on the applied quantization method.
\begin{table}[H]
    \centering
    \setlength{\arrayrulewidth}{0.4mm}
    \begin{tabular}{|l|c|c|c|c|}
        \hline
        \textbf{Model} & \textbf{Blocks} & \textbf{raw} & \textbf{8bit} & \textbf{4bit} \\
        \hline
        meta-llama/Meta-Llama-3.1-8B-Instruct & 32 & 0.4062 & 0.2031 & 0.1079 \\
        Qwen/Qwen2-7B-Instruct & 28 & 0.4341 & 0.2171 & 0.1153 \\
        google/gemma-2-9b-it & 42 & 0.3692 & 0.1846 & 0.0981 \\
        microsoft/Phi-3.5-mini-instruct & 32 & 0.2109 & 0.1055 & 0.0560 \\
        \hline
    \end{tabular}
    \vspace{5px}
    \caption{Comparison of transformer model block sizes across different quantization types}
    \label{tab:transformer_block_sizes}
\end{table}

\subsubsection{Classifiers Comparison}\label{sec_calassifiers_comparison}

To compare the performance of two Fast classifiers based on the results obtained in the FastEWQ Test Results section, we introduce a composite score formula that combines accuracy and perplexity values from the MMLU benchmark. Since accuracy values range from 0 to 1, while perplexity values are generally larger, we apply the natural logarithm to the perplexity values to bring them to a comparable scale.

The composite score is computed using weights for perplexity (\(w_1\)) and accuracy (\(w_2\)), both of which are set to 1, indicating that we value both metrics equally when calculating the score. The composite score is given by
\[
\text{Composite Score} = w_1 \cdot \log(\text{Perplexity}) - w_2 \cdot \text{Accuracy}
\]
The inputs for calculating the composite score for each model are provided in the table below:

\begin{table}[H]
    \centering
    \setlength{\arrayrulewidth}{0.4mm}
    {
    \begin{tabular}{| l | l | l |}
        \hline
        \textbf{Variant} & \textbf{Accuracy} & \textbf{Perplexity} \\
        \hline
        fast 8bit mixed & 0.6826, 0.6894, 0.6461, 0.6238 & 2.2379, 3.1906, 4.3702, 4.104 \\
        fast 4bit/8bit mixed & 0.6833, 0.688, 0.6458, 0.6253 & 2.2332, 3.2203, 4.2577, 4.0964 \\
        fast train 8bit mixed & 0.6822, 0.6876, 0.647, 0.6238 & 2.2379, 3.1827, 4.3397, 4.0879 \\
        fast train 4bit/8bit mixed & 0.6824, 0.6875, 0.6453, 0.6246 & 2.2325, 3.2126, 4.2561, 4.1334 \\
    \hline
    \end{tabular}}
    \vspace{5px}
    \caption{Inputs for Composite Score Calculation (collected from Table \ref{tab:model_performance_fast})}
    \label{tab:composite_score}
\end{table}
To compare the performance of the classifiers, we observe three different combinations of classifiers, including a trained FastEWQ classifier on the entire dataset (fast) and a trained FastEWQ classifier on 70\% of the samples (fast train). For comparison metrics, we use the paired \(t\)-test and Cohen's \(d\).

\textbf{Paired \(t\)-test} is a statistical method used to compare the means of two related groups to determine if there is a significant difference between them. This test is useful when comparing two sets of measurements taken from the same group or sample, such as performance scores of the same classifier under different conditions. The null hypothesis assumes that there is no significant difference between the paired values.

The test statistic is calculated as
\[
t = \frac{\bar{d}}{s_d / \sqrt{n}}
\]
where \(\bar{d}\) is the mean of the differences between paired observations, \(s_d\) being the standard deviation of the differences, and \(n\) is the number of paired observations. The \(p\)-value is then obtained to determine whether the difference is statistically significant.

\begin{table}[H]
\centering
\setlength{\arrayrulewidth}{0.4mm}
\begin{tabular}{|c|c|}
\hline
\textbf{\(p\)-value} & \textbf{Significance} \\
\hline
\( p < 0.05 \) & significant \\
\hline
\( 0.05 \leq p < 0.10 \) & marginally significant \\
\hline
\( p \geq 0.10 \) & not significant \\
\hline
\end{tabular}
\vspace{5px}
\caption{Significance levels for \(p\)-values}
\label{tab:pvalue_significance}
\end{table}

\textbf{Cohen's \(d\)} is a measure of the effect size, or the magnitude of the difference between two groups. It is commonly used to quantify the size of the difference in means relative to the variability observed in the data. Cohen's \(d\) is calculated as
\[
d = \frac{\bar{X_1} - \bar{X_2}}{s_p}
\]
where \(\bar{X_1}\) and \(\bar{X_2}\) are the sample means, and \(s_p\) is the pooled standard deviation, which combines the standard deviations of the two groups. Cohen's \(d\) helps to understand not just whether a difference exists (like the \(p\)-value) but how large that difference is.

\begin{table}[H]
\centering
\setlength{\arrayrulewidth}{0.4mm}
\begin{tabular}{|c|c|}
\hline
\textbf{Cohen's \(d\) Value} & \textbf{Effect Size Interpretation} \\
\hline
\( d < 0.2 \) & negligible \\
\hline
\( 0.2 \leq d < 0.5 \) & small \\
\hline
\( 0.5 \leq d < 0.8 \) & medium \\
\hline
\( d > 0.8 \) & large \\
\hline
\end{tabular}
\vspace{5px}
\caption{Interpretation of Cohen's \(d\) values}
\label{tab:cohen_effect_size}
\end{table}

The results of the paired classifier comparison are shown in the Figure \ref{fig:composite_score_comparison} and Table \ref{tab:classifiers_comparison} below.

\begin{figure}[H]
\centering
\includegraphics[width=0.95\textwidth]{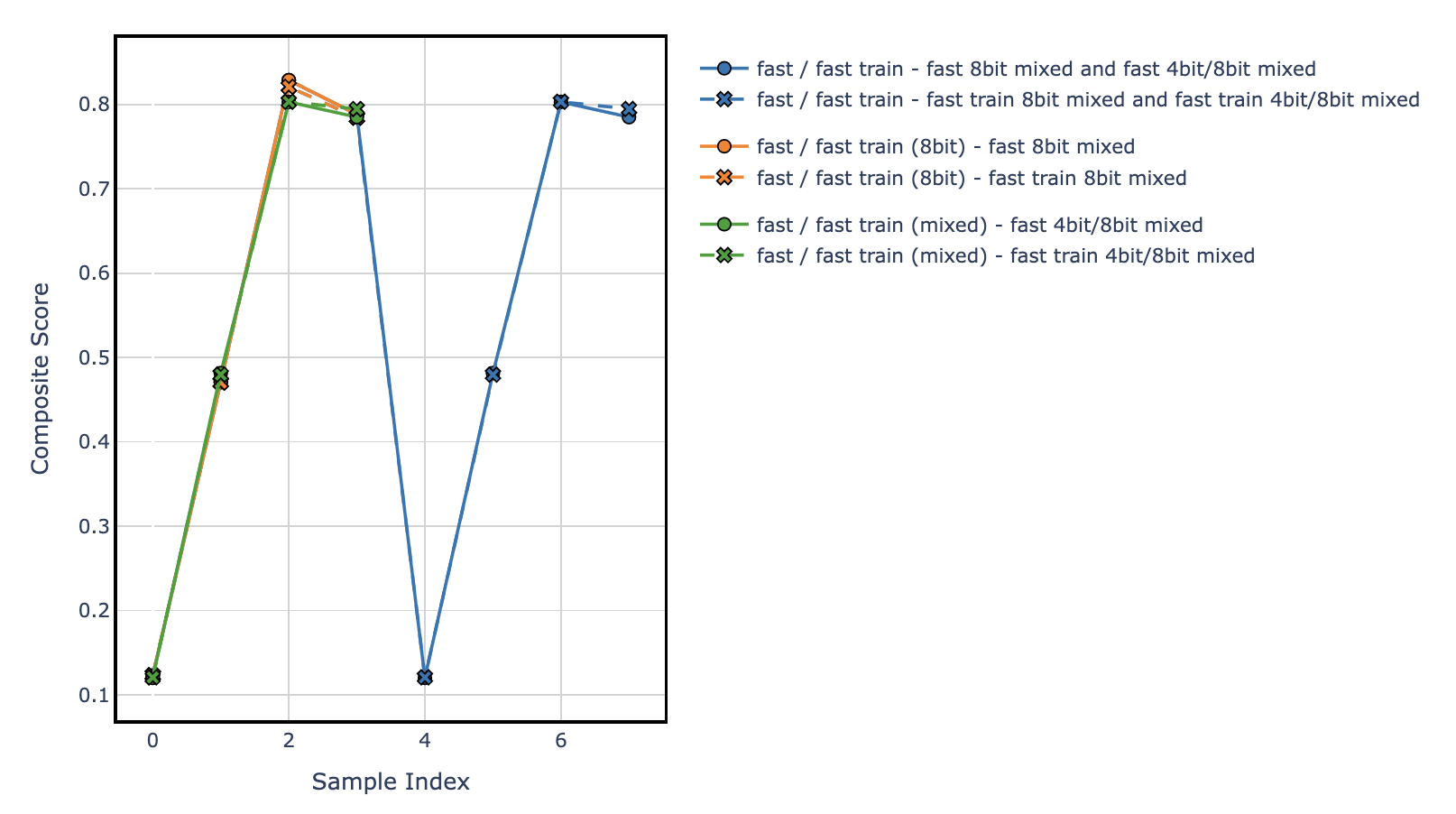}
\caption{Comparison of composite scores between classifiers}
\label{fig:composite_score_comparison}
\end{figure}

\begin{table}[H]
    \centering
    \setlength{\arrayrulewidth}{0.4mm}
    \resizebox{\textwidth}{!}{
    \begin{tabular}{| l | l | l | l | l | l |}
        \hline
        \textbf{Comparison} & \textbf{Variant} & \textbf{Abs Diff} & \textbf{\(t\)-statistic} & \textbf{\(p\)-value / Effect} & \textbf{Cohen's \(d\) / Effect} \\
        \hline
        fast / fast train & fast 8bit mixed and fast 4bit/8bit mixed & 0.0031 & 0.2551 & 0.806 / not significant & 0.0016 / not significant \\
        fast / fast train & fast train 8bit mixed and fast train 4bit/8bit mixed & {} & {} & not significant & not significant \\
        \hline
        fast / fast train (8bit) & fast 8bit mixed & 0.0032 & 1.6215 & 0.2034 / not significant & 0.0107 / not significant \\
        fast / fast train (8bit) & fast train 8bit mixed & {} & {} & not significant & not significant \\
        \hline
        fast / fast train (mixed) & fast 4bit/8bit mixed & 0.0031 & -0.825 & 0.4699 / not significant & -0.0076 / not significant \\
        fast / fast train (mixed) & fast train 4bit/8bit mixed & {} & {} & not significant & not significant \\
        \hline
    \end{tabular}}
    \vspace{5px}
    \caption{Statistical comparison of composite scores between classifiers}
    \label{tab:classifiers_comparison}
\end{table}
The paired classifier comparison results, as shown in Table \ref{tab:classifiers_comparison}, summarize the outcomes of statistical tests performed on various classifier variants. The key metrics considered are the {Abs Diff}, {\(t\)-statistic}, {\(p\)-value}, and {Cohen's \(d\)}. The Abs Diff metric represents the mean absolute difference between composite scores, as shown in Figure \ref{fig:composite_score_comparison}. We observe that the value for classifier pairs is very small, approximately 0.0031, and from the figure, we observe that the composite scores nearly overlap between classifiers, indicating a negligible difference when comparing them.

Based on the statistical analysis, we conclude that there is no significant difference in performance between the trained classifier variants under the tested conditions. This implies that a classifier trained on 70\% of the training set effectively captures the behavior of one trained on the entire dataset. Since the trained classifier is not overfitted, it demonstrates promising generalization properties on unseen model architectures, as the underlying concept is retained. The \(t\)-statistic values indicate relatively small differences, and the corresponding \(p\)-values confirm that none of these differences are statistically significant. Furthermore, Cohen's \( d \) values suggest negligible effect sizes, reinforcing the conclusion that the classifiers perform similarly across different configurations.

\subsection{Behavior Capture and Summary}

In Table~\ref{tab:model_performance_comparison}, we present the final results, highlighting relative differences in accuracy, perplexity, model size, and EWQ analysis time complexity.

Based on the test results, we conclude that the EWQ method significantly enhances the performance of the \texttt{meta-llama/Meta-Llama-3.1-8B-Instruct} model. When applying the EWQ 4-bit/8-bit mixed quantization optimization, the model achieves lower perplexity than its unquantized counterpart while maintaining comparable accuracy, indicating improved coherence and overall performance.

For the \texttt{Qwen/Qwen2-7B-Instruct}, FastEWQ 8-bit mixed quantization achieves the highest accuracy. The minimal perplexity difference between the quantized and unquantized models suggests that quantization does not substantially impact performance.

In the case of \texttt{google/gemma-2-9b-it}, the unquantized version performs best. However, the optimal trade-off between accuracy, perplexity, and space efficiency is achieved through 8-bit or 8-bit mixed quantization of the transformer blocks, allowing the model to fit on devices with 16GB of memory. The best-performing model is obtained using the FastEWQ classifier trained on a subset of the training dataset, demonstrating strong generalization. This model selects three additional transformer blocks for quantization, and with strict quantization, it achieves the best perplexity and accuracy for the 8-bit quantization strategy while reducing model size by 19.77\% compared to the unquantized version.

Notable results are observed with Microsoft’s Phi model, \texttt{microsoft/Phi-3.5-mini-instruct}. The quantized models perform nearly identically to their unquantized counterparts. Specifically, a FastEWQ classifier trained on a subset of the training dataset achieves almost the same accuracy and perplexity as the unquantized model while reducing its size by 18\%.

The FastEWQ classifier trained on a subset of the training dataset generally performs as well as or better than an overfitted version. A comparison between the two reveals no significant differences, only minor variations, indicating that the subset-trained FastEWQ classifier generalizes better to unseen models. For 8-bit quantization, FastEWQ effectively replicates EWQ’s results while offering $O(1)$ time complexity for analysis, compared to $O(n)$ for standard EWQ. Additionally, FastEWQ outperforms global quantization, which represents only a subset of the possible combinations FastEWQ can generate, making global quantization a special case of FastEWQ.

In conclusion, considering the trade-offs between accuracy, perplexity, model size, time complexity, generalization on unseen models, and real-time analysis support, the optimal choice is the FastEWQ optimizer trained on a subset of the training data.

\begin{table}[H]
    \centering
    \setlength{\arrayrulewidth}{0.4mm}
    \resizebox{\textwidth}{!}{
    \begin{tabular}{|l|l|c|c|c|c|}
        \hline
        \textbf{Model} & \textbf{Variant} & \textbf{Accuracy} & \textbf{Perplexity} & \textbf{Size / Total (GB)} & \textbf{Complexity} \\
        \hline
        meta-llama/Meta-Llama-3.1-8B-Instruct & raw & 0.6837 & 2.2379 & 16.07 & - \\
        meta-llama/Meta-Llama-3.1-8B-Instruct & 4bit & -3.2\% & 5.02\% & -71.87\% / 4.52 & O(1) \\
        meta-llama/Meta-Llama-3.1-8B-Instruct & 8bit & -0.47\% & 0.01\% & -46.92\% / 8.53 & O(1) \\
        meta-llama/Meta-Llama-3.1-8B-Instruct & 8bit mixed & -0.25\% & -0.03\% & -17.8\% / 13.21 & O(n) \\
        meta-llama/Meta-Llama-3.1-8B-Instruct & 4bit/8bit mixed & -0.22\% & -0.33\% & -18.98\% / 13.02 & O(n) \\
        meta-llama/Meta-Llama-3.1-8B-Instruct & fast 8bit mixed & -0.16\% & 0.0\% & -17.8\% / 13.21 & O(1) \\
        meta-llama/Meta-Llama-3.1-8B-Instruct & fast 4bit/8bit mixed & -0.06\% & -0.21\% & -18.29\% / 13.13 & O(1) \\
        meta-llama/Meta-Llama-3.1-8B-Instruct & fast train 8bit mixed & -0.22\% & 0.0\% & -17.8\% / 13.21 & O(1) \\
        meta-llama/Meta-Llama-3.1-8B-Instruct & fast train 4bit/8bit mixed & -0.19\% & -0.24\% & -18.29\% / 13.13 & O(1) \\
        \hline
        Qwen/Qwen2-7B-Instruct & raw & 0.6872 & 3.1722 & 15.23 & - \\
        Qwen/Qwen2-7B-Instruct & 4bit & -1.99\% & 5.7\% & -62.9\% / 5.65 & O(1) \\
        Qwen/Qwen2-7B-Instruct & 8bit & -0.51\% & 0.56\% & -43.01\% / 8.68 & O(1) \\
        Qwen/Qwen2-7B-Instruct & 8bit mixed & 0.32\% & 0.58\% & -20.16\% / 12.16 & O(n) \\
        Qwen/Qwen2-7B-Instruct & 4bit/8bit mixed & 0.04\% & 1.92\% & -22.32\% / 11.83 & O(n) \\
        Qwen/Qwen2-7B-Instruct & fast 8bit mixed & 0.32\% & 0.58\% & -20.16\% / 12.16 & O(1) \\
        Qwen/Qwen2-7B-Instruct & fast 4bit/8bit mixed & 0.12\% & 1.52\% & -21.01\% / 12.03 & O(1) \\
        Qwen/Qwen2-7B-Instruct & fast train 8bit mixed & 0.06\% & 0.33\% & -18.71\% / 12.38 & O(1) \\
        Qwen/Qwen2-7B-Instruct & fast train 4bit/8bit mixed & 0.04\% & 1.27\% & -19.37\% / 12.28 & O(1) \\
        \hline
        google/gemma-2-9b-it & raw & 0.6505 & 4.1013 & 18.41 & - \\
        google/gemma-2-9b-it & 4bit & -3.4\% & 52.57\% & -66.11\% / 6.24 & O(1) \\
        google/gemma-2-9b-it & 8bit & -0.86\% & 5.42\% & -48.61\% / 9.46 & O(1) \\
        google/gemma-2-9b-it & 8bit mixed & -0.68\% & 6.56\% & -18.36\% / 15.03 & O(n) \\
        google/gemma-2-9b-it & 4bit/8bit mixed & -0.52\% & 11.66\% & -21.18\% / 14.51 & O(n) \\
        google/gemma-2-9b-it & fast 8bit mixed & -0.68\% & 6.56\% & -18.36\% / 15.03 & O(1) \\
        google/gemma-2-9b-it & fast 4bit/8bit mixed & -0.72\% & 3.81\% & -18.85\% / 14.94 & O(1) \\
        google/gemma-2-9b-it & fast train 8bit mixed & -0.54\% & 5.81\% & -19.77\% / 14.77 & O(1) \\
        google/gemma-2-9b-it & fast train 4bit/8bit mixed & -0.8\% & 3.77\% & -20.26\% / 14.68 & O(1) \\
        \hline
        microsoft/Phi-3.5-mini-instruct & raw & 0.6243 & 4.0805 & 7.62 & - \\
        microsoft/Phi-3.5-mini-instruct & 4bit & 0.14\% & 11.32\% & -69.69\% / 2.31 & O(1) \\
        microsoft/Phi-3.5-mini-instruct & 8bit & -0.29\% & 0.33\% & -47.38\% / 4.01 & O(1) \\
        microsoft/Phi-3.5-mini-instruct & 8bit mixed & -0.08\% & 0.58\% & -23.75\% / 5.81 & O(n) \\
        microsoft/Phi-3.5-mini-instruct & 4bit/8bit mixed & -0.75\% & 3.23\% & -26.38\% / 5.61 & O(n) \\
        microsoft/Phi-3.5-mini-instruct & fast 8bit mixed & -0.08\% & 0.58\% & -23.75\% / 5.81 & O(1) \\
        microsoft/Phi-3.5-mini-instruct & fast 4bit/8bit mixed & 0.16\% & 0.39\% & -24.41\% / 5.76 & O(1) \\
        microsoft/Phi-3.5-mini-instruct & fast train 8bit mixed & -0.08\% & 0.18\% & -18.24\% / 6.23 & O(1) \\
        microsoft/Phi-3.5-mini-instruct & fast train 4bit/8bit mixed & 0.05\% & 1.3\% & -18.9\% / 6.18 & O(1) \\
        \hline
    \end{tabular}}
    \vspace{5px}
    \caption{MMLU performance vs. model size across quantization methods}
    \label{tab:model_performance_comparison}
\end{table}

\subsection{Further Analysis of FastEWQ Optimization}

Expanding on the results discussed in Table~\ref{tab:model_performance_comparison}, we further analyze the behavior of the FastEWQ optimizer trained on the full dataset. Notably, for 8-bit quantization, FastEWQ closely approximates the selections made by the EWQ method, reinforcing its effectiveness as a lightweight alternative. This alignment is particularly relevant given the role of 8-bit quantization in matrix decomposition for attention computations, as seen in LLM.int8() \citep{dettmers2022llmint88bitmatrixmultiplication}. 

For models such as \texttt{Qwen/Qwen2-7B-Instruct}, \texttt{google/gemma-2-9b-it}, and \texttt{microsoft/Phi-3.5-mini-instruct}, FastEWQ identifies nearly the same transformer blocks for quantization as EWQ, reflecting prior observations that efficient classifiers can approximate computationally expensive sensitivity analyses \citep{Li2021BRECQPT}. However, for \texttt{meta-llama/Meta-Llama-3.1-8B-Instruct}, architectural variations result in minor deviations in block selection. Despite these differences, the impact on performance remains negligible ($\sim 6 \times 10^{-4}$), aligning with findings that minor quantization discrepancies are often absorbed by model redundancy \citep{xiao2024smoothquant}. 

Beyond accuracy and perplexity, FastEWQ’s $O(1)$ analysis time complexity provides a substantial speedup over iterative EWQ, with at least a 100x efficiency gain. Notably, for \texttt{google/gemma-2-9b-it}, training FastEWQ on 70\% of the dataset outperforms the full-dataset variant by selecting three additional blocks for quantization, leading to a 19.77\% reduction in model size. This finding supports previous research \citep{ashkboos2024quarot}, which highlights how subset training mitigates overfitting in quantization controllers. 

In comparing 8-bit mixed and fast 8-bit mixed quantization, FastEWQ consistently reproduces EWQ’s behavior across most models, further validating its reliability. The flexibility of this method surpasses that of global quantization, representing only a single point in the broader mixed-precision strategy space \citep{Gong2019MixedPN}. Consequently, FastEWQ offers an optimal balance between accuracy, perplexity, model size, and computational efficiency, reinforcing its role as the preferred quantization strategy.

\section{Conclusion}  
We introduced Entropy-Weighted Quantization (EWQ), a novel architecture- and size-agnostic method for post-training quantization of LLMs. By analyzing entropy distributions across transformer blocks, EWQ identifies layers amenable to precision reduction while preserving critical high-entropy components. Our experiments demonstrate that EWQ maintains MMLU accuracy within 0.5\% of full-precision models while reducing memory usage by up to 18\%—outperforming uniform quantization baselines like GPTQ \citep{frantar2022gptq}—and achieves superior perplexity in some cases, a phenomenon attributed to quantization-induced regularization. The method’s effectiveness spans diverse architectures (1.6B to 70B parameters), including LLaMA, Qwen, Phi, and Gemma, proving its universality across model scales and designs.  

FastEWQ, an optimized variant, eliminates weight-loading requirements through a classifier that predicts quantization suitability using execution index, parameter count, and total blocks. This approach achieves 80\% classification accuracy with \(O(1)\) time complexity, enabling real-time deployment decisions. Both EWQ and FastEWQ significantly enhance the feasibility of running state-of-the-art LLMs on resource-constrained devices, such as 16GB consumer hardware, without performance degradation.

\subsection{Future Directions}  
As we advance toward more efficient and adaptable quantization techniques, several promising research directions emerge. The evolution of model compression must balance precision, computational efficiency, and hardware compatibility, ensuring that quantization techniques remain both effective and scalable. In this context, we identify three key areas for further exploration.
\begin{itemize}  
    \item[1.] \textbf{Architectural Generalization}: Adapting EWQ principles to non-transformer architectures (e.g., SSMs \citep{gu2023mamba}, RWKV \citep{peng2023rwkv}) and multimodal models.  
    \item[2.] \textbf{Precision Frontiers}: Exploring sub-4-bit quantization (2-bit, 1.58-bit \citep{ashkboos2023quik}) combined with entropy-aware sparsity, building on sparsity-aware methods like SparseGPT \citep{frantar2023sparsegpt}.  
    \item[3.] \textbf{System Integration}: Co-designing EWQ with emerging memory technologies (HBM3, CXL) and kernel-level optimizations, inspired by FlashAttention \citep{dao2022flashattention} and vLLM \citep{kwon2023efficientmemorymanagementlarge}.  
\end{itemize}  

Additional opportunities include theoretical investigations into entropy-robustness relationships, extending information-theoretic frameworks like \citep{shwartzziv2017openingblackboxdeep}, and federated learning applications where FastEWQ’s metadata-driven approach could enable dynamic precision allocation across distributed systems \citep{kairouz2021advancesopenproblemsfederated}. Finally, integrating EWQ with activation quantization \citep{yao2022zeroquant} and KV cache compression \citep{liu2025llmsmaintainfundamentalabilities} could unlock end-to-end efficient inference pipelines for next-generation LLMs.  

\vspace{0.8cm}
\begin{spacing}{0.1} 
\footnotesize 
\begin{multicols}{2}
\bibliography{ewq}
\end{multicols}
\end{spacing}

\end{document}